\documentclass[letterpaper, 10 pt, conference]{ieeeconf}  %

\IEEEoverridecommandlockouts                              %

\usepackage{graphics} %
\usepackage{epsfig} %
\usepackage{mathptmx} %
\usepackage{times} %
\usepackage{amsmath} %
\usepackage{amssymb}  %

\usepackage{multirow} %

\usepackage{cite}
\usepackage{atbegshi}

\usepackage{paralist}
\usepackage{url}
\usepackage{lipsum}
\usepackage{graphicx}
\usepackage[locale=US]{siunitx}

\makeatletter
\let\MYcaption\@makecaption
\makeatother

\usepackage[font=footnotesize]{subcaption}

\makeatletter
\let\@makecaption\MYcaption
\makeatother

\usepackage{fontawesome}

\usepackage{tikz}
\usetikzlibrary{arrows}
\usetikzlibrary{shapes}
\usetikzlibrary{snakes}
\usetikzlibrary{calc}
\usetikzlibrary{decorations.markings, shadows, spy, positioning}
\usepackage{pgf}
\usepackage{pgfplots}
\usepgfplotslibrary{units}
\usepgfplotslibrary{groupplots,dateplot}
\usetikzlibrary{patterns,shapes.arrows}
\pgfplotsset{compat=newest}
\usetikzlibrary{tikzmark,calc}
\usetikzlibrary{arrows}
\usetikzlibrary{shapes}
\usetikzlibrary{snakes}
\usetikzlibrary{calc}
\usetikzlibrary{arrows}
\usetikzlibrary{shapes}
\usetikzlibrary{snakes}
\usetikzlibrary{calc}
\usetikzlibrary{tikzmark,fit}
\usetikzlibrary{positioning}
\usetikzlibrary {arrows.meta}
\usetikzlibrary{spy}
\usepackage{pgfplots}
\usetikzlibrary{pgfplots.statistics, pgfplots.colorbrewer, pgfplots.groupplots}
\usepackage{pgfplotstable}
\usepackage{tikz-3dplot}

\usepackage{amsmath}
\usepackage{amssymb}

\makeatletter
\let\NAT@parse\undefined
\makeatother
\usepackage[hidelinks,backref=section]{hyperref}

\usepackage{etoolbox}

\renewcommand*{\backref}[1]{}
\renewcommand*{\backrefalt}[4]{%
    \ifcase #1 %
    \or
        (Sec.~#2)%
    \else
        (Secs.~#2)%
    \fi
}

\DeclareMathOperator*{\argmax}{arg\,max}

\newcommand{\videolink}{at \url{https://youtu.be/_d7AqTRjz6g}}
\newcommand{\codelink}{\url{https://github.com/wheelbot/mini-wheelbot}}

\title{\LARGE \bf
The Mini Wheelbot: A Testbed for Learning-based \\ Balancing, Flips, and Articulated Driving
}

\author{Henrik Hose, Jan Weisgerber, and Sebastian Trimpe%
\thanks{This work is funded in part by the German Research Foundation (DFG) – RTG 2236/2 (UnRAVeL). Simulations were performed with computing resources granted by
RWTH Aachen University under project RWTH1621.}%
\thanks{All authors are with the Institute for Data Science in Mechanical Engineering (DSME), RWTH Aachen University, Germany,
        {\tt\small \{henrik.hose, trimpe\}@dsme.rwth-aachen.de}}%
}

\AtBeginShipoutNext{%
  \AtBeginShipoutUpperLeft{%
    \begin{tikzpicture}[remember picture, overlay]
      \node[anchor=south,yshift=0.5cm,text width=\textwidth,align=justify,font=\scriptsize] at (current page.south) {
        Accepted final version. Accepted for publication in: \textit{2025 IEEE International Conference on Robotics and Automation (ICRA)}, 2025. DOI: 10.1109/ICRA55743.2025.11128020.\\
        © 2025 IEEE. Personal use of this material is permitted. Permission from IEEE must be obtained for all other uses, in any current or future media, including reprinting/republishing this material for advertising or promotional purposes, creating new collective works, for resale or redistribution to servers or lists, or reuse of any copyrighted component of this work in other works.
      };
    \end{tikzpicture}%
  }%
}

\begin{document}

\maketitle
\thispagestyle{empty}
\pagestyle{empty}

\begin{abstract}
The Mini Wheelbot is a balancing, reaction wheel unicycle robot designed as a testbed for learning-based control.
It is an unstable system with highly nonlinear yaw dynamics, non-holonomic driving, and discrete contact switches in a small, powerful, and rugged form factor.
The Mini Wheelbot can use its wheels to stand up from any initial orientation -- enabling automatic environment resets in repetitive experiments and even challenging half flips.
We illustrate the effectiveness of the Mini Wheelbot as a testbed by implementing two popular learning-based control algorithms.
First, we showcase Bayesian optimization for tuning the balancing controller.
Second, we use imitation learning from an expert nonlinear MPC that uses gyroscopic effects to reorient the robot and can track higher-level velocity and orientation commands.
The latter allows the robot to drive around based on user commands -- for the first time in this class of robots.
The Mini Wheelbot is not only compelling for testing learning-based control algorithms, but it is also just fun to work with, as demonstrated in the video of our experiments \videolink.

\end{abstract}

\section{Introduction}
Experimental validation is an integral part of robotics research that even grows in importance with data-driven and learning-based control relying on real-world data.
To test novel algorithms, researchers have proposed a multitude of balancing and driving robots in recent years~\cite{geist2022wheelbot, hofer2023one, carron2023chronos, bodmer2024optimization, o2020f1tenth}.
However, driving robots lack the challenge of instability, and pure balancing lacks mobility and thus higher-level tasks, as we will detail in the discussion of related work.
Despite many practical robots having nonlinear, unstable, and hybrid system behavior (e.g.,~quadcopters~\cite{hanover2024autonomous} and~legged robots~\cite{kim2017design}), small-scale test systems with all these properties are rare.
We think that validating learning algorithms in the presence of uncertainty requires i) safety, ii) robustness~(i.e., not break easily), and iii) automatic experimentation~(i.e.,~reset  after failure), all of which we aim to achieve.

\begin{figure}[t]
    \centering
    \includegraphics[width=3.4in,trim={3.4cm 0.3cm 0cm 0.9cm},clip]{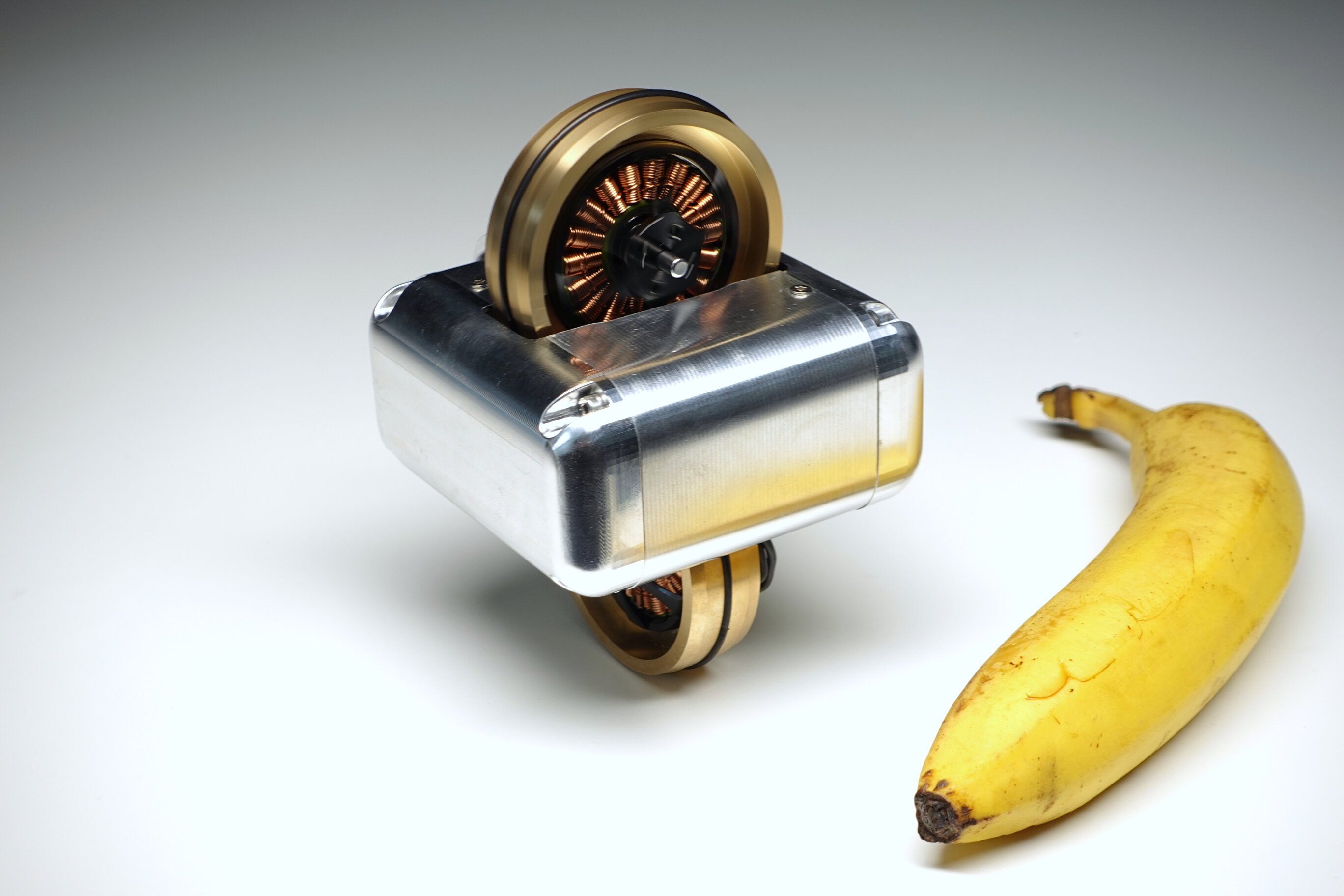}%
    \caption{The Mini Wheelbot: A small, rugged, and symmetric reaction wheel unicycle robot with challenging nonlinear, unstable, and hybrid dynamics. The Mini Wheelbot can stand up from any position which allows for automatic environment resets in learning-based control experiments.}\label{fig:firstfig}
    \vspace{-2mm}
\end{figure}

The Mini Wheelbot is a symmetric, reaction wheel, balancing unicycle robot with directly controlled, unstable roll and pitch dynamics.
The yaw state is uncontrollable for the linearized system.
This necessitates nonlinear methods utilizing gyroscopic effects to achieve articulated (meaningful) driving.
The ability to stand up from any initial position enables environment resets for automated experiments.
Additionally, the Mini Wheelbot is powerful enough for interesting maneuvers like flips.
These episodic and continuing tasks are abstractions of typical learning problems in robotics.
In this paper, we showcase two learning-based algorithms to solve them.
First, we use Bayesian optimization~(BO) to tune a balancing controller with minimal human intervention due to automatic resets.
Second, we approximate a sophisticated model-predictive controller~(MPC) with a fast-to-evaluate neural network to control the yaw orientation and drive around based on keyboard commands.

The Mini Wheelbot is a complete re-design that substantially advances an early prototype in~\cite{geist2022wheelbot}: It is smaller, more robust, and has a powerful CPU running Linux while being affordable with bill of material costs of~\$1200 per robot (at quantity ten).
In summary, we make the following contributions:
\begin{compactenum}
    \item The Mini Wheelbot, a small, powerful, rugged, open-source robot ideal for learning-based control experiments due to automatic environment resets and interesting dynamics.
    \item Implementation of two state-of-the-art learning approaches (BO for controller tuning, approximate MPC) to illustrate the versatility as a learning testbed.
    \item With the approximate MPC, we achieve yaw control and thus articulated (meaningful) driving for the first time for this type of robot.
\end{compactenum}
A video of the Mini Wheelbot and our experiments is available~\videolink.

\section{Related Work}
Small-scale driving~\cite{mondada1994mobile,johnson2006mobile,mondada2009puck,rubenstein2015aerobot,pickem2015gritsbot,wilson2016pheeno, paull2017duckietown, hsieh2006economical, liniger2015optimization, carron2023chronos, bodmer2024optimization, goldfain2019autorally, o2020f1tenth, gonzales2018planning} and balancing~\cite{gajamohan2012cubli, mayr2015mechatronic, muehlebach2016nonlinear, hofer2023one, klemm2019ascento, nagarajan2014ballbot, geist2022wheelbot} robots are a popular choice for control experiments, though none is specifically designed for testing learning algorithms.
This section summarizes existing designs, highlighting that either articulated driving or balancing of an unstable system is achieved.
In contrast, the Mini Wheelbot combines both with highly nonlinear yaw dynamics and automatic environment resets ideal for learning experiments.

\subsection{Driving Robots}
Driving robots have been developed for decades, commonly with slow and stable differential drives~\cite{mondada1994mobile,johnson2006mobile,mondada2009puck,rubenstein2015aerobot,pickem2015gritsbot,wilson2016pheeno, paull2017duckietown} or fast and car-like dynamics for autonomous racing~\cite{hsieh2006economical, liniger2015optimization, carron2023chronos, bodmer2024optimization, goldfain2019autorally, o2020f1tenth, gonzales2018planning}.
Differential-drive platforms are cheap to produce in large quantities for testing networked and multi-robot control algorithms~\cite{johnson2006mobile, paull2017duckietown, wilson2016pheeno,pickem2015gritsbot, pickem2017robotarium, schwab2020experimental}, or as education systems~\cite{mondada1994mobile, mondada2009puck, rubenstein2015aerobot}.
Research questions arise mainly from high-level perception and coordination while simple kinematic planning and state-feedback controllers perform well in steering.
On car-like robots, pushing the limits of autonomous racing requires precise control of fast dynamics and difficult-to-model tire slip effects, however, the underlying contouring control problem can be linearized and is inherently stable when driving slowly, hence even PID schemes succeed in this task~\cite{carron2023chronos}.
In comparison, the Mini Wheelbot is unstable and has interesting nonlinear tasks (yaw control) that can not be achieved by classic linear methods.

\subsection{Balancing Robots}
Balancing is a long-standing challenge in robotic research, for example in pendulum sculptures~\cite{trimpe2012balancing,gajamohan2012cubli, mayr2015mechatronic,hofer2023one}, unicycle robots~\cite{geist2022wheelbot, schoonwinkel1988design, vos1990dynamics, xu2011pendulum, daud2017dynamic, lee2012decoupled, jae2011fuzzy, li2012attitude, rosyidi2016speed, neves2021discrete, rizal2015point, jin2010balancing} like the Mini Wheelbot, and even legged~\cite{klemm2019ascento} or ball~\cite{nagarajan2014ballbot} robots.
Cube-like robots stabilize standing on a corner with leavers~\cite{trimpe2012balancing} or reaction wheels~\cite{gajamohan2012cubli,mayr2015mechatronic,hofer2023one}, where nonlinearity~\cite{muehlebach2016nonlinear} and underactuation~\cite{hofer2023one} inspire research.
While some cube-like robots can automatically stand up~\cite{gajamohan2012cubli, mayr2015mechatronic, muehlebach2016nonlinear}, their mobility is restricted to walking-like sequences of controlled stand-up and falling.
Similarly, the Mini Wheelbot can use its reaction wheel to balance on the point contact of the driving wheel, stand up from any initial position, and has underactuated yaw dynamics.
That is, when in perfect balance, the Mini Wheelbot has no direct control over it's yaw angle.
Compared to cube-like robots, however, the driving wheel allows for mobility.

Balancing unicycle robots use levers~\cite{schoonwinkel1988design,jin2010balancing,xu2011pendulum,daud2017dynamic} or reaction wheels~\cite{geist2022wheelbot,neves2021discrete,jae2011fuzzy, rosyidi2016speed,lee2012decoupled,rizal2015point}.
Except for an early prototype~\cite{geist2022wheelbot}, existing designs' actuators lack power for a stand-up and are asymmetric, which prohibits interesting chaining of stand-up maneuvers.
Roll and pitch balancing with linear state-feedback or fuzzy controllers is well understood~\cite{xu2011pendulum, jae2011fuzzy, lee2012decoupled, neves2021discrete}, also with reference velocities~\cite{rosyidi2016speed}.
However, so far, a third turntable actuator was required to control the yaw orientation~\cite{schoonwinkel1988design,vos1990dynamics,rizal2015point,jin2010balancing} and thus permit meaningful driving.
In contrast, we use nonlinear methods to control yaw without an additional turntable actuator for the first time.

\section{Design of the Mini Wheelbot}
The Mini Wheelbot needs to be small, rigid, and powerful for learning experiments.
In this section, we summarize the custom design of hardware and electronics that is necessary to achieve these design goals and meet the specifications listed in Tab.~\ref{tab:robot_specs}.
\begin{table}[hbt]
    \caption{Specifications of the Mini Wheelbot.}
    \centering
    \begin{tabular}{|l|l|}
    \hline
    \textbf{Spec.} & \textbf{Value} \\
    \hline
    \hline
    Dims. & \SI{130}{\milli\meter} (height) x \SI{87}{\milli\meter} (width) \\
    \hline
    Weight & Total \SI{0.69}{\kilo\gram}, wheels \SI{0.13}{\kilo\gram} (each), body \SI{0.43}{\kilo\gram} \\
    \hline
    Motors & T-Motor MN4006, max.~\num{8000}~rpm, \SI{0.5}{\newton\meter} \\
    \hline
    Battery & \SI{450}{\milli\ampere\hour}, \SI{22.2}{\volt}, 6S LiPo, max. cont. \SI{20}{\ampere} discharge  \\
    \hline
    Runtime & \SI{45}{\minute} when balancing\\
    \hline
    CPU & Pi CM4, BCM2711 quad-core Cortex-A72, \SI{1.5}{\giga\hertz} \\
    \hline
    Sensors & 4x Bosch BMI088 IMU, 2x AMS AS5047D 14bit encoder\\
    \hline
    System & RT-Preempt Linux, Buildroot \\
    \hline
    \end{tabular}
    \label{tab:robot_specs}
\end{table}

\subsection{Mechanical Design}

\begin{figure}[tb]
    \centering
    \input{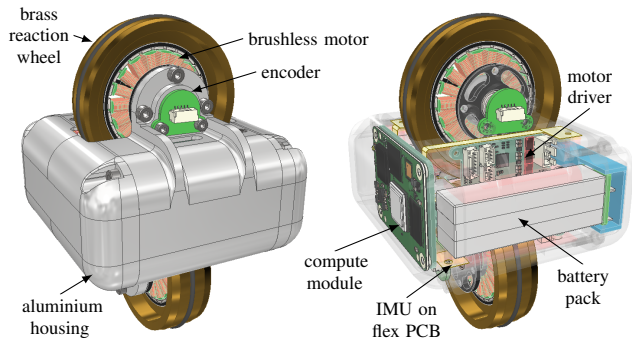}
    \caption{System overview of the Mini Wheelbot.}
    \label{fig:packaging}
\end{figure}

The Mini Wheelbot is designed to be compact, rugged, and powerful.
Its size is determined by the onboard compute module (a Raspberry Pi CM4) and six battery cells.
Brass is chosen as material for the reaction wheels due to its high density and good availability at CNC job shops.
Sizing calculations of the reaction wheels were performed as described in~\cite{geist2022wheelbot} to maximize rotational inertia needed for the stand-up maneuvers.
The body is made from aluminum, which is lightweight, durable, and inexpensive to machine.
Inside the body, compute module, battery pack, and power electronics are packed tightly (see Fig.~\ref{fig:packaging}).
The packaging minimizes body mass and inertia for highly dynamic maneuvers like stand-up and flips while keeping symmetry.

\subsection{Electronics Design}
\begin{figure}[b]
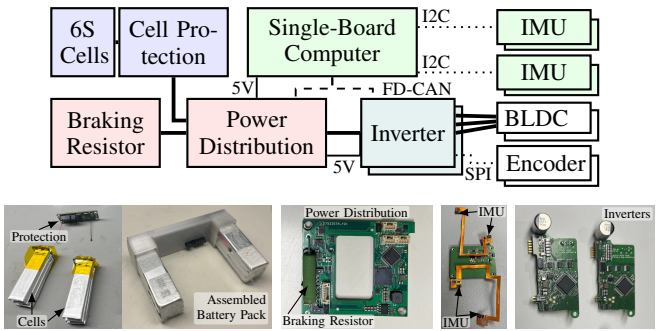

    \centering
    \begin{subfigure}[b]{3.4in}
        \centering
        \begin{tikzpicture}[scale=1]   

    \node [rectangle, thick, draw, align=center, fill=blue!10, minimum height = 0.9cm, text width=0.6cm, font={\small}] (Cells) at (-1.5,1.2) 
    {6S Cells};

    \node [rectangle, thick, draw, align=center, fill=blue!10, minimum height = 0.9cm, text width=1.2cm, font={\small}] (BatteryProt) at (-0.3,1.2) 
    {Cell Protection};

    \node [rectangle, thick, draw, align=center, fill=red!10, minimum height = 0.9cm, text width=1.6cm, font={\small}] (PowerDistro) at (0.8,0)
    {Power Distribution};
    \node [rectangle, thick, draw, align=center, fill=red!10, minimum height = 0.9cm, text width=1.2cm, font={\small}] (BrakingResistor) at (-1.2,0)
    {Braking Resistor};
    
    \node [rectangle, thick, draw, align=center, fill=teal!10, minimum height = 0.9cm, text width=1cm, font={\small}] (InverterShadow) at (2.9,-0.1){};
    \node [rectangle, thick, draw, align=center, fill=teal!10, minimum height = 0.9cm, text width=1cm, font={\small}] (Inverter) at (2.8,0)
    {Inverter};

    \node [rectangle, thick, draw, align=center, fill=white, minimum height = 0.4cm, text width=1cm, align=center, font={\small}] (BLDCShadow) at (4.7,0.1)
    {};
    \node [rectangle, thick, draw, align=center, fill=white, minimum height = 0.4cm, text width=1cm, align=center, font={\small}] (BLDC) at (4.6,0.2)
    {};
    \node[align=center, font={\small}] at (4.5, 0.2) {BLDC};

    \node [rectangle, thick, draw, align=center, fill=white, minimum height = 0.4cm, text width=1cm, align=center, font={\small}] (EncoderShadow) at (4.7,-0.5)
    {};
    \node [rectangle, thick, draw, align=center, fill=white, minimum height = 0.4cm, text width=1cm, align=center, font={\small}] (Encoder) at (4.6,-0.4)
    {Encoder};

    \node [rectangle, thick, draw, align=center, fill=green!10, minimum height = 0.9cm, text width=2cm, font={\small}] (SBC) at (1.8,1.2)
    {Single-Board Computer};
    
    \node [rectangle, thick, draw, align=center, fill=green!10, minimum height = 0.4cm, text width=1cm, font={\small}] (IMU1Shadow) at (4.7,0.7)
    {};
    \node [rectangle, thick, draw, align=center, fill=green!10, minimum height = 0.4cm, text width=1cm, font={\small}] (IMU1) at (4.6,0.8)
    {IMU};
    
    \node [rectangle, thick, draw, align=center, fill=green!10, minimum height = 0.4cm, text width=1cm, font={\small}] (IMU2Shadow) at (4.7,1.3)
    {};
    \node [rectangle, thick, draw, align=center, fill=green!10, minimum height = 0.4cm, text width=1cm, font={\small}] (IMU2) at (4.6,1.4)
    {IMU};

    \begin{scope}[
        every node/.style={rectangle, align=center, font={\scriptsize}},
        every edge/.style={thick, draw, line width = 0.25mm},
        every path/.style={thick, draw, line width = 0.25mm}
    ]

        \draw[dotted] (SBC.east)++(0,-0.4) |- node [above right=-0.1cm]{I2C} (IMU1.west);
        \draw[dotted] (SBC.east)++(0,0.2)  |- node [above right=-0.1cm]{I2C} (IMU2.west);
        \draw[dotted] (Inverter.-25) -| ++(0.2,0) |- (Encoder.west) node [below left=-0.1cm]{SPI} ;
        \draw[-] (SBC.south)++(-1,0) -- node [left=-0.1cm]{5V} (PowerDistro.north);
        \draw[dashed] (SBC.south)++(0,0) -- ++(0,-0.15) -| node [right]{FD-CAN} (Inverter.135);
        \draw[dashed] (SBC.south)++(0,0) -- ++(0,-0.15) -| (PowerDistro.42);
        \draw[-] (PowerDistro.east)++(0,-0.3) -| node [below left=-0.1cm]{5V}  (Inverter.west);
    \end{scope}
    
    \begin{scope}[
        every node/.style={rectangle, align=center, font={\scriptsize}},
        every edge/.style={very thick, draw, line width = 0.5mm},
        every path/.style={very thick, draw, line width = 0.5mm}
    ]
        \draw[-] (Cells) -- (BatteryProt);
        \draw[-] (BatteryProt.south) |- (PowerDistro.170);
        \draw[-] (PowerDistro.west) -- (BrakingResistor.east);
        \draw[-] (PowerDistro) -- (Inverter);
        \draw[-] (Inverter.20) --  (BLDC);
        \draw[-] (Inverter.10)  --  (BLDC);
        \draw[-] (Inverter.0) -- (BLDC);
    \end{scope}

\end{tikzpicture}
    \end{subfigure}
    \par\medskip %
    \begin{subfigure}[b]{3.4in}
        \centering
        \newcommand{\photoheight}{0.64in}
        \input{figures/electronics/photos/battery.tex}%
        \input{figures/electronics/photos/power_distribution.tex}%
        \input{figures/electronics/photos/compute_imu.tex}%
        \input{figures/electronics/photos/inverter.tex}%
    \end{subfigure}
    \caption{Electronics design of the Mini Wheelbot (top) and custom circuit boards (bottom).}
    \label{fig:electronicsschematic}
\end{figure}

The custom electronics inside the Mini Wheelbot are designed to maximize the robots power and control performance while ensuring safety and battery runtime.
An overview of custom electronics is shown in Fig.~\ref{fig:electronicsschematic}.
Schematics for all components are available online\footnote{\label{footnote:code}\codelink}.

The removable battery pack has six \SI{450}{\milli\ampere\hour} Lithium Polymer cells (three on each side of the robot) protected by two BQ77915 with overcurrent and short-circuit protection -- all safely potted in one housing (Fig.~\ref{fig:electronicsschematic}, bottom left).
The power distribution board monitores power consumption and engages a braking resistor in case recuperated current during stand-up maneuvers can not be charged back into battery or external power supply.
In addition, a~nRF24L01P communicates with a wireless emergency stop button to shut off motor controllers.
Two custom inverters based on the $\mu$Motor~\cite{lehmann2021micro} run field-oriented control (FOC) to drive T-Motor MN4006 brushless DC motors.
Current is measured low-side on all phases at FOC loop frequency of~\SI{40}{\kilo\hertz}.
Due to careful analog design and INA241A2 current sense amplifiers, the gains of the PI d-q-current controllers can be chosen as high as~\SI{10}{\kilo\hertz} bandwidth.
On-axis magnetic 14-bit AS5047 encoders are interfaced by the inverters via SPI with transfers triggered from an interrupt synchronized with the FOC loop.
The motors exhibit strong cogging with peaks of up to~\SI{20}{\milli\newton\meter} (\SI{4}{\percent} of the maximum torque).
This severely impacts balancing control.
We therefore implement cogging compensation with a lookup table of feedforward torques.
These are calibrated by sweeping through a full rotation in \num{4000} steps with a high-gain position controller and saving the average current to hold each position in flash memory on the inverter (cf.~\cite{piccoli2016anticogging} for details).
The compensation effectively reduces the cogging to less than~\SI{2.5}{\milli\newton\meter} (\SI{0.5}{\percent}).
Power distribution and inverters are commanded via CAN-FD with an MCP2517FD controller by the Raspberry Pi CM4 compute module at~\SI{1}{\kilo\hertz}.
Bosch BMI088 IMUs are directly interfaced via I2C from the CM4.

\subsection{Software}
The Mini Wheelbot runs a Buildroot Linux with real-time kernel.
The single-board computer performs all higher-level estimation and control, implemented in C++ and running inside a Docker container.
While sending CAN messages through Linux SocketCAN is almost latency free in this setup, there is significant delay (up to~\SI{10}{\milli\second}) between the kernel interrupt and receiving encoder values on the user-space socket.
This delay is due to the processing of received CAN messages by the same workers as all other networking and can be compensated in software by extrapolation.
IMUs readings are available through the Linux IIO subsystem.

\section{A Versatile Control Testbed}

\begin{figure}[tb]
    \centering
    \input{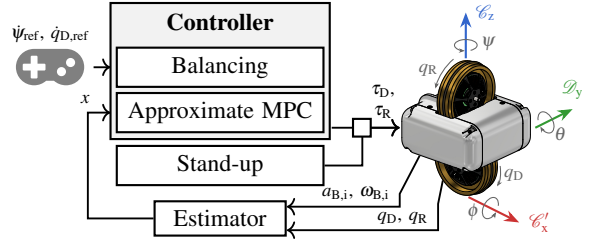}
    \caption{Control system overview and generalized coordinates.}
    \label{fig:controlsystem}
\end{figure}

This section details the control methods on the Mini Wheelbot, as visualized in~Fig.~\ref{fig:controlsystem}, that we later use in illustrative learning tasks in Sec.~\ref{sec:lbcexperiments}.
To this end, we first describe the system model (Sec.~\ref{sec:systemmodel}) and optimization-based identification of parameters from trajectory data (Sec.~\ref{sec:parameteridentification}), before introducing 
the state-feedback controller used for balancing (Sec.~\ref{sec:statefeedback}), and a nonlinear model-predictive controller for the orientation of the robot that is required for meaningful driving (Sec.~\ref{sec:nonlinearmpc}).
Finally, the stand-up in the roll and pitch direction and a half-flip are described (Sec.~\ref{sec:standup} and \ref{sec:flip}).
All implementations are available online\footnotemark[1].

We omit a detailed description on the estimation of the robots state, as we implement the existing method from~\cite{geist2022wheelbot} to fuse four gyroscope\footnote{Using corrected gyroscope measurement equations from the errata of~\cite{geist2022wheelbot}.}~$\omega_\text{B,i}$, four accelerometer~$a_\text{B,i}$, and two wheel encoder~$q_\text{D,R}$ measurements~(cf.~\cite{trimpe2010accelerometer,trimpe2012balancing,gajamohan2012cubli,muehlebach2016nonlinear} for details). Notably, the presented controllers are robust to alignment errors of the IMUs thus no calibration is needed.

\subsection{System Model}
\label{sec:systemmodel}
The Mini Wheelbot's state can be described as~$x~=~[\psi,\phi,\theta,\dot{\psi},\dot{\phi},\dot{\theta},q_\text{D}, q_\text{R}, \dot{q}_\text{D}, \dot{q}_\text{R}]^\top$.
The body orientation is expressed in Euler-like angles in yaw-roll-pitch order (see Fig.~\ref{fig:controlsystem}) with yaw~$\psi$ and roll~$\phi$ around the contact point coordinate system's~$\mathcal{C}_\text{z}$ axis and resulting~$\mathcal{C}^\prime_\text{x}$ axis, and pitch~$\theta$ around the drive motors axis of rotation that is aligned with~$\mathcal{D}_\text{y}$.
The angles of the wheels are~$q_\text{D,R}$ and the actions~$u~=~[\tau_\text{D}, \tau_\text{R}]^\top$ are the torques applied to driving~(bottom) and the reaction~(top) wheel.
The Mini Wheelbot's nonlinear, continuous-time dynamics of implicit form~$f_\text{ct}(x,\dot{x}, u, p)=0$ with parameters~$p$ can be derived using standard, multi-body methods~(e.g.,  Euler-Lagrange equations).
A detailed symbolic derivation using a computer algebra system is given in the supplementary code\footnotemark[1] and described in detail in~\cite{daud2017dynamic}.
The resulting differential equations have the form:
\begin{align}\label{eqn:eqnofmotion}
    \begin{split}M(\phi,\theta,p)&[\ddot{\phi},\ddot{\theta},\ddot{\psi}, \ddot{q}_\text{D}, \ddot{q}_\text{R}]^\top + b(\phi, \theta, \dot{\psi}, \dot{\phi}, \dot{\theta}, \dot{q}_\text{D}, \dot{q}_\text{R},p) \\ &+ g(\phi, \theta, \tau_\text{D}, \tau_\text{R},p)+\tau_{\psi}(\dot{\psi},p)= 0.
    \end{split}
\end{align}
The implicit dynamics~(\ref{eqn:eqnofmotion}) are parameterized by $p~=~[m_\text{D,R}, m_{B}, I_{\text{D}_\text{x,z}\text{,R}_\text{y,z}}, I_{\text{D}_\text{y}\text{,R}_\text{x}},
I_\text{B},
r_\text{D,R},
l_\text{D,R},
C_1, C_2]\in\mathbb{R}^{11}$, which consists of the masses and mass moments of inertia of the robot's body~$m_\text{B}$ and $I_\text{B}$ and reaction wheels~$m_\text{D,R}$ and ~$I_\text{D,R}$.
The mass moments of inertia are assumed to be diagonal and share values for multiple axis due to~(quasi-)symmetries.
Further, the radius of the wheels~$r_\text{D,R}$ and distance between rotation axis~$l_\text{D,R}$ enter as geometric parameters.
The rotational friction in the contact point is modeled as~$\tau_{\psi}=C_1\tanh(C_2\dot{\psi})$, where the constant~$C_1$ models the magnitude and~$C_2$ the slope of the friction.
The mass matrix~$M(\phi,\theta)$ is difficult to invert symbolically (i.e., symbolically computing~$\dot{x}=f_\text{expl}(x,u,p)$) and (\ref{eqn:eqnofmotion}) is stiff on a timescale relevant for controlling the yaw orientation.
Both issues motivate the use of implicit integrators~\cite{frey2023fast} for an accurate discrete-time model with time~$t\in\mathbb{N}$, that is:
\begin{align} \label{eqn:systemdynamics}
        x(t+1) = f_\text{dt}(x(t), u(t), p).
\end{align}

\subsection{Optimization-based Parameter Identification}
\label{sec:parameteridentification}
In this section, the optimization-based identification of the parameters~$p$ from data is described, which is similar to the methods presented in~\cite{bock1983recent,valluru2017development,simpson2023efficient,bodmer2024optimization}.
To this end, an optimization problem
\begin{align}
\begin{split}\label{eqn:sysid}
\min_{p\in\mathcal{P}, \hat{x},\hat{u}} & \sum_{i=0}^{N_\text{id}} \sum_{t=0}^{T_\text{id}}\|\hat{x}_i(t)-\bar{x}_i(t)\|^2_{Q_\text{id}}+\|\hat{u}_i(t)-\bar{u}_i(t)\|^2_{R_\text{id}}\\
\text{s.t.}\quad 
&\hat{u}_i(t)-\bar{u}_i(t)\in\mathcal{W},\;\forall t\in\mathbb{I}_{T_\text{id}},\forall i\in\mathbb{I}_{N_\text{id}}\\
&\hat{x}_i(t+1) = f_\text{dt}(\hat{x}(t), \hat{u}_i(t),p)
\;\forall t\in\mathbb{I}_{T_\text{id}},\forall i\in\mathbb{I}_{N_\text{id}}\\
\end{split}
\end{align}
is solved, where the parameters are constrained to reasonable box constraints~$\mathcal{P}$, the optimization objective consists of quadratic cost on the error between predicted~$\hat{x}$ and measured state~$\bar{x}$ and predicted~$\hat{u}$ and measured action~$\bar{u}$ over a finite horizon~$T_\text{id}$ and a set of trajectories~$N_\text{id}$.
The input disturbance constraint set~$\mathcal{W}$ can be conservatively approximated from direct measurements of the actuator torque ripple.
Quadratic cost matrices are chosen empirically as diagonal matrices~$Q_\text{id}=\text{diag}([100,1,1,10,10,10,0.1,0.1,0.1,0.1]^\top)$ and~$R_\text{id}=\text{diag}([100,100]^\top)$.

\begin{figure}[thpb]
    \centering
    \input{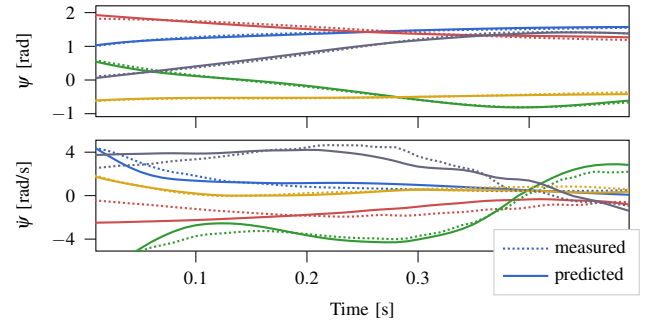}
    \caption{Five of the trajectories used for parameter identification: measurements (dotted) of yaw angle (top) and yaw rotational velocity (bottom) and predictions with optimal parameters~$p^*$ (solid).}
    \label{fig:sysid}
\end{figure}

The optimization problem (\ref{eqn:sysid}) is implemented in CasADi~\cite{andersson2019casadi} and solved with IPOPT~\cite{wachter2006implementation}, which yields the optimal parameters~$p^*$.
A selection of trajectories from the dataset and the achieved prediction for the critical yaw state is plotted in Fig.~\ref{fig:sysid}.
In the following, the discrete-time dynamics~(\ref{eqn:systemdynamics}) with~$p^*$ substituted in will be referred to as~$f_\text{dt}(x(t), u(t))$, dropping the explicit dependency on~$p$ for ease of notation.

\subsection{Balancing State-Feedback Controller}
\label{sec:statefeedback}
The Mini Wheelbot can balance in the roll and pitch direction using a simple state-feedback controller as proposed in~\cite{geist2022wheelbot}.
While we use the same structure, that is, state-feedback for the roll and the pitch states separately, we choose a data-driven approach to tune the gains of the controllers directly from closed-loop experiments instead of LQR (see Sec.~\ref{sec:bo}).

The state feedback controller~$u(t) = Kx(t)$ has the feedback gains matrix
\begin{align}
    \setlength\arraycolsep{3pt}
    K = \begin{bmatrix}
        0 &  0  & K_\text{D1} & 0 & 0   & K_\text{D2}   & K_\text{D3} & 0   & K_\text{D4} & 0 \\
        0 & K_\text{R1} & 0   & 0 & K_\text{R2} & 0   & 0 & K_\text{R3} & 0   & K_\text{R4}
    \end{bmatrix},
    \label{eqn:feedbackgain}
\end{align}
where the gains~$K_\text{D}\in\mathbb{R}^4$ control pitch and~$K_\text{R}\in\mathbb{R}^4$ roll.

\textit{Remark}: The yaw orientation of the robot is non-controllable for the Mini Wheelbot's dynamics linearized about the equilibrium balancing position, which are~$A=\tfrac{\partial f_\text{expl}}{\partial x} |_{x=0,u=0}$ and $B=\tfrac{\partial f_\text{expl}}{\partial u}|_{x=0,u=0}$.
The controllability matrix~$C=[B, AB, A^2B,\dots]$ has~$\text{rank}(C)=9 < 10$ where the non-controllable state is $\psi$.
Yet, friction in the wheel-to-ground contact lets~$\dot{\psi}$ converge to standstill.

\subsection{Nonlinear MPC for Driving Control}
\label{sec:nonlinearmpc}
Controlling the Mini Wheelbot's yaw orientation is required for driving the robot along meaningful paths, but can not be achieved by simple, linear control methods.
We propose using a nonlinear MPC~\cite{rawlings2017model} for controlling the yaw orientation.
In MPC, an optimization problem is repeatedly solved to compute a sequence of optimal actions that minimize a cost function for predicted future states.
From the sequence of optimal actions, only the first is applied in closed loop before solving the optimization problem again.
For the Mini Wheelbot, we formulate the nonlinear MPC
\begin{subequations}\label{eqn:mpc}
\begin{align}
\min_{v, x} & \sum_{k=0}^{N_\text{MPC}} \|x(t|k)\|^2_{Q_\text{MPC}} + \|u(t|k)\|^2_{R_\text{MPC}} \label{eqn:mpc:cost}\\
\text{s.t.}\quad & x(t|k+1) = f_\text{dt}(x(t|k), u(t|k)), \label{eqn:mpc:dyn}\\
& u(t|k) = K x(t|k) + v(t|k), \label{eqn:mpc:prestab}\\ %
& x(t|k)\in\mathcal{X}, u(t|k)\in\mathcal{U} \; \forall k\in\mathbb{I}_{N_\text{MPC}-1}\label{eqn:mpc:constr}\\
& x(t|0) = x(t),\quad x(t|N_\text{MPC}) \in \mathcal{X}_f \label{eqn:mpc:initterm}
\end{align}
\end{subequations}
that includes pre-stabilizing feedback (\ref{eqn:mpc:prestab}) with the controller from Sec.~\ref{sec:statefeedback} and a terminal constraint (\ref{eqn:mpc:initterm}).
The terminal constraint is~$\mathcal{X}_f=0$ to guarantee convergence of the solver to a solution that drives the yaw angle to the desired setpoint.
After solving the optimization problem~(\ref{eqn:mpc}), the first action is applied to the system,~$u(t)=u^*(t|0)$.
Cost terms are empirically chosen as diagonal matrices~$R_\text{MPC} = \text{diag}([10, 0.01]^\top)$ and~$Q_\text{MPC} = \text{diag}([100,1,1,0.001, 0.01, 1, 1, 0.0001, 0.25, 0.001]^\top)$.
Details on the constraints are available in the supplementary code\footnotemark[1].
We implement the nonlinear MPC optimization problem~(\ref{eqn:mpc}) in CasADi~\cite{andersson2019casadi} and solve it with IPOPT~\cite{wachter2006implementation}.

\subsection{Stand-Up Maneuver}

\begin{figure}[tb]
    \begin{subfigure}{1.6in}
        \includegraphics[height=0.5in,trim={3cm  3cm 37cm 6cm},clip]{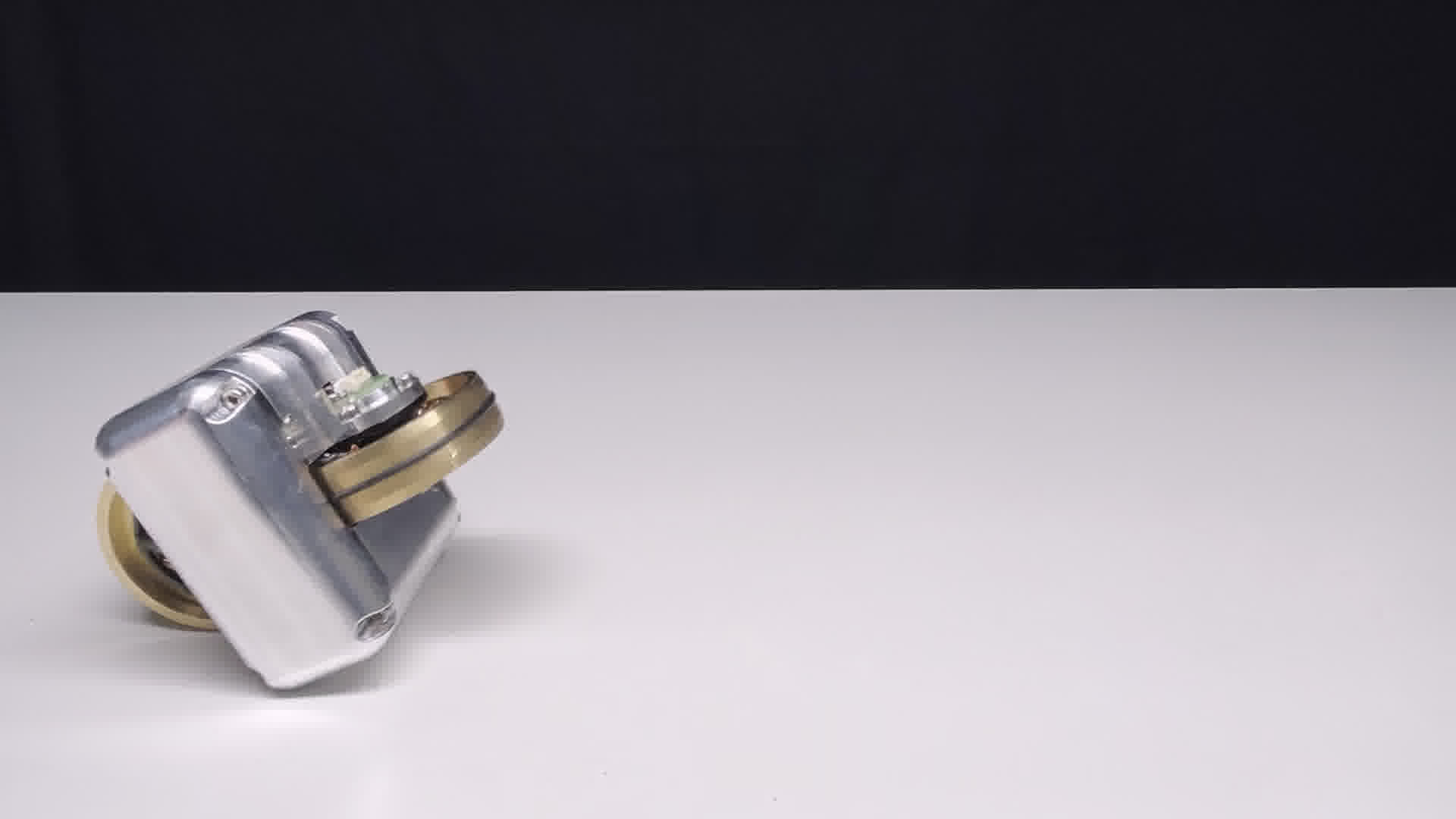}%
        \includegraphics[height=0.5in,trim={23cm 3cm 15cm 6cm},clip]{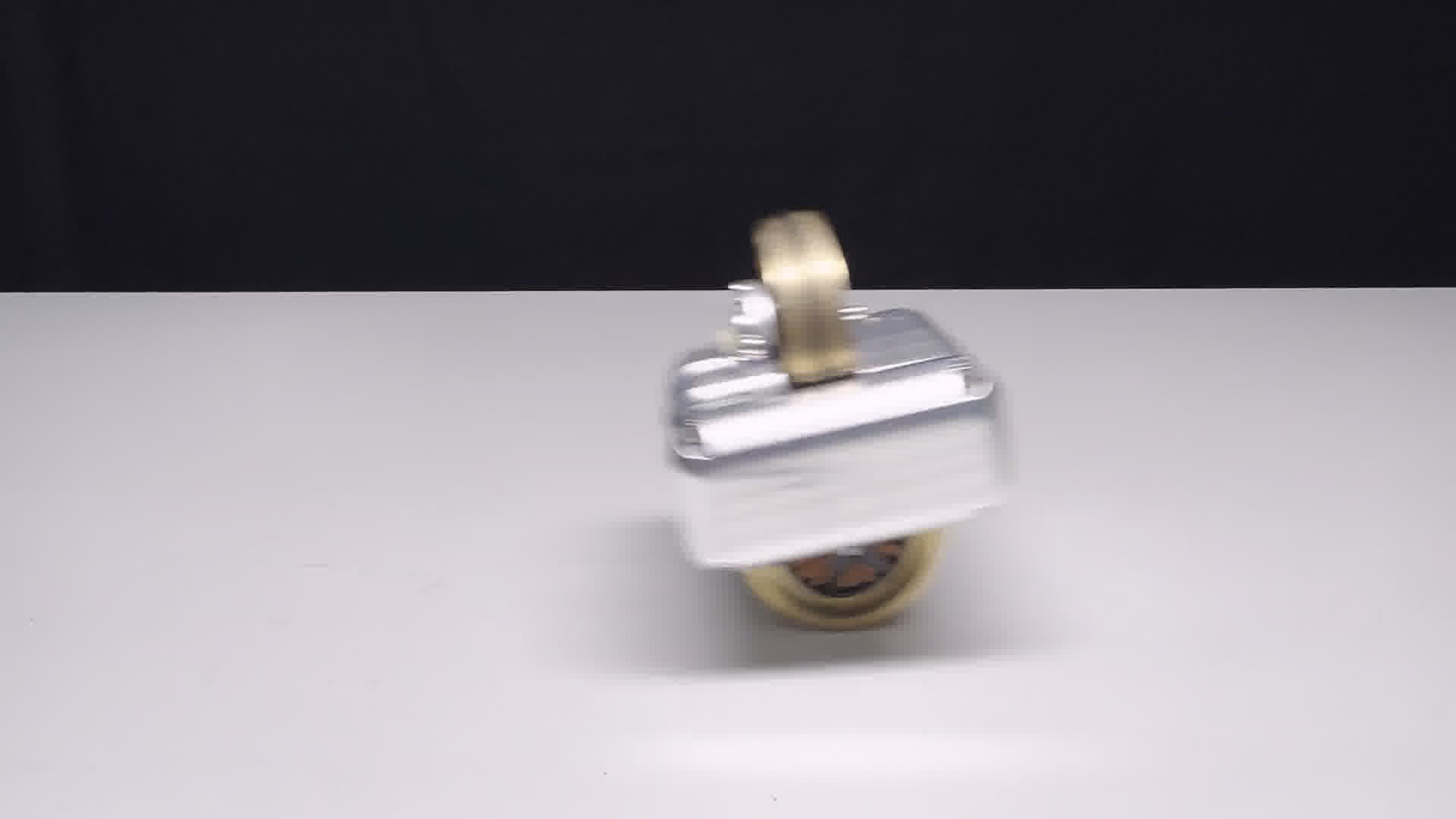}%
        \includegraphics[height=0.5in,trim={35cm 3cm  0cm 6cm},clip]{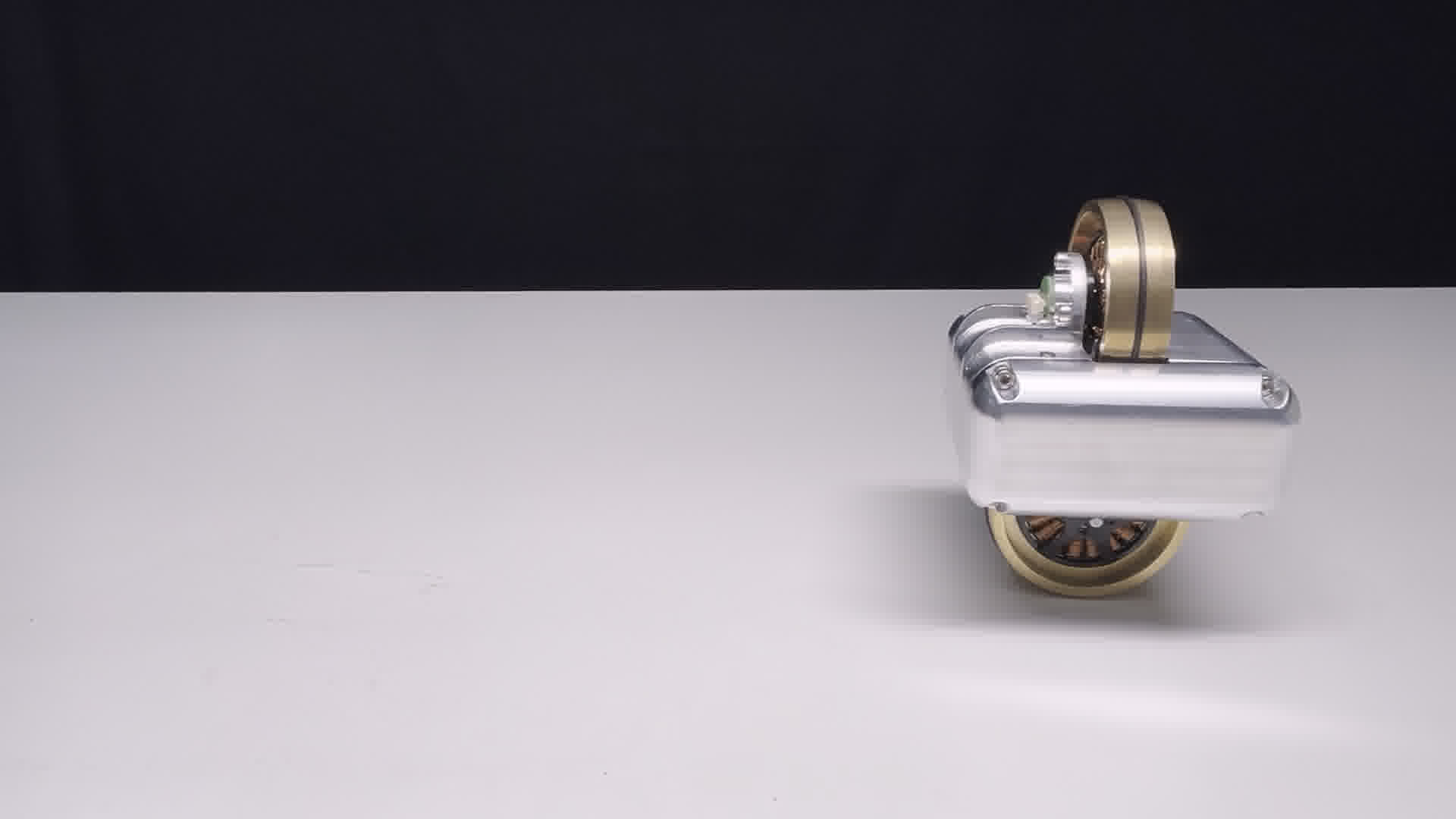}%
    \end{subfigure}
    \begin{subfigure}{1.6in}
        \includegraphics[height=0.5in,trim={0cm  7cm 35cm 0cm},clip]{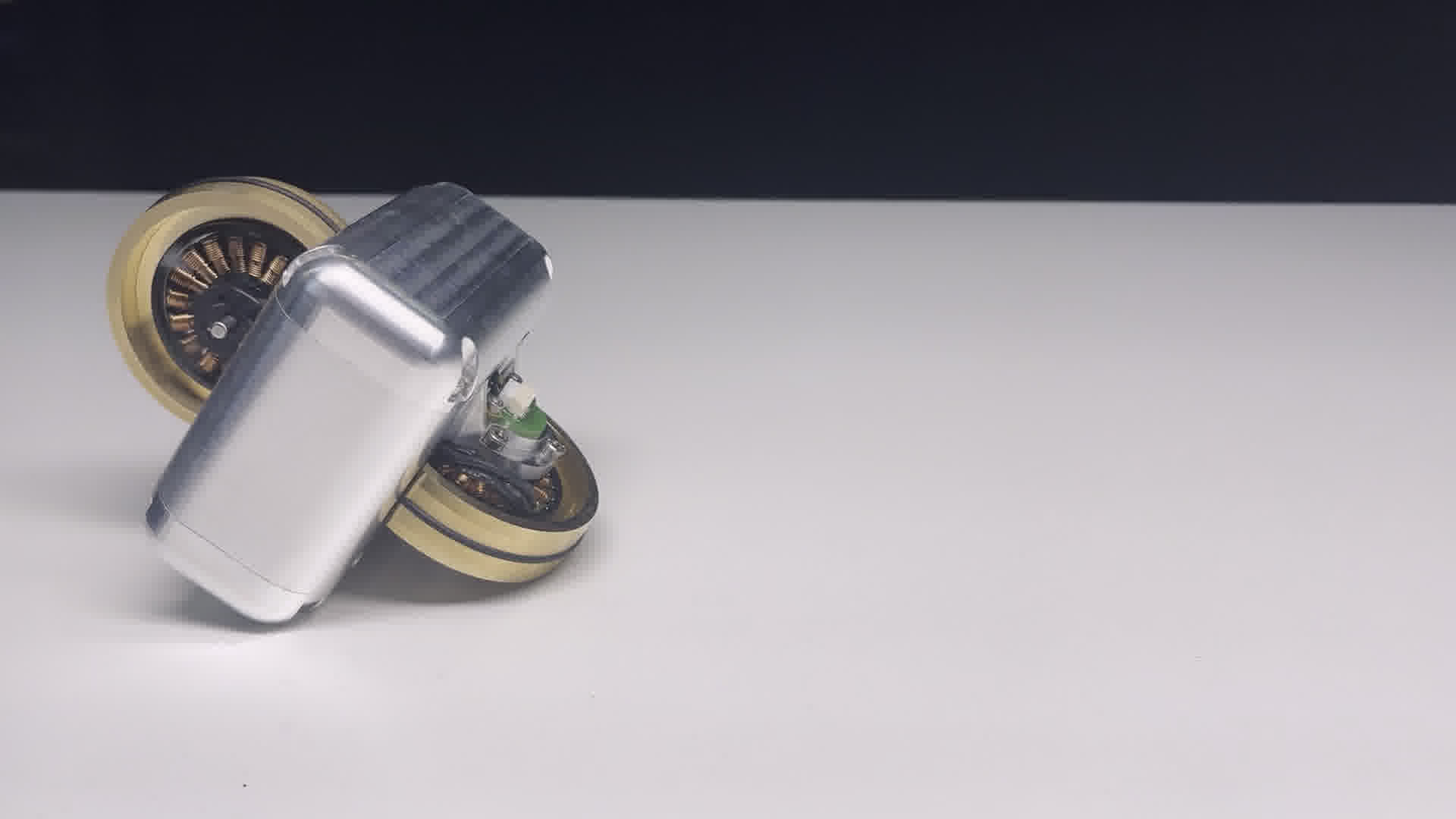}%
        \includegraphics[height=0.5in,trim={5cm  7cm 33cm 0cm},clip]{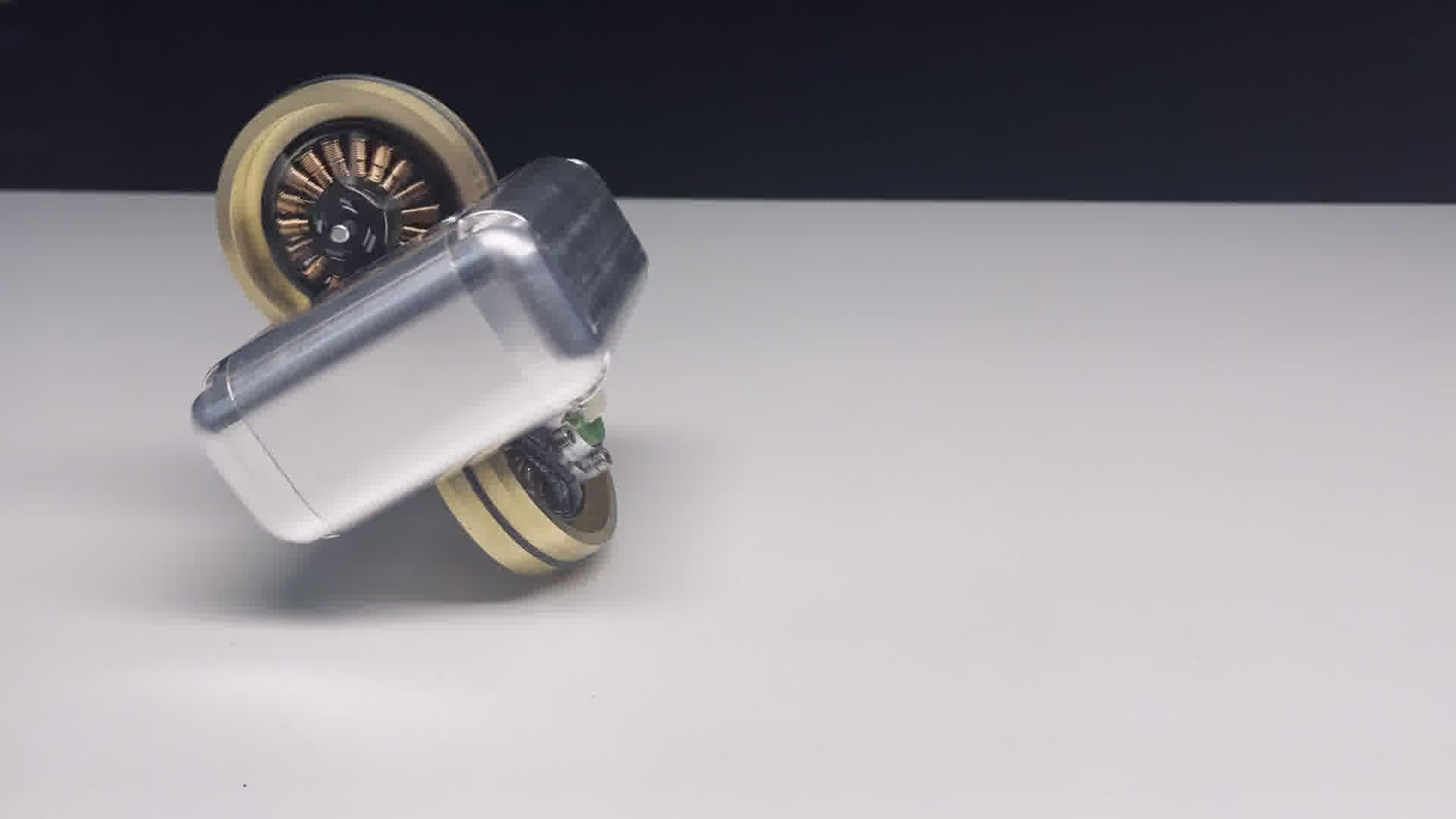}%
        \includegraphics[height=0.5in,trim={15cm 7cm 25cm 0cm},clip]{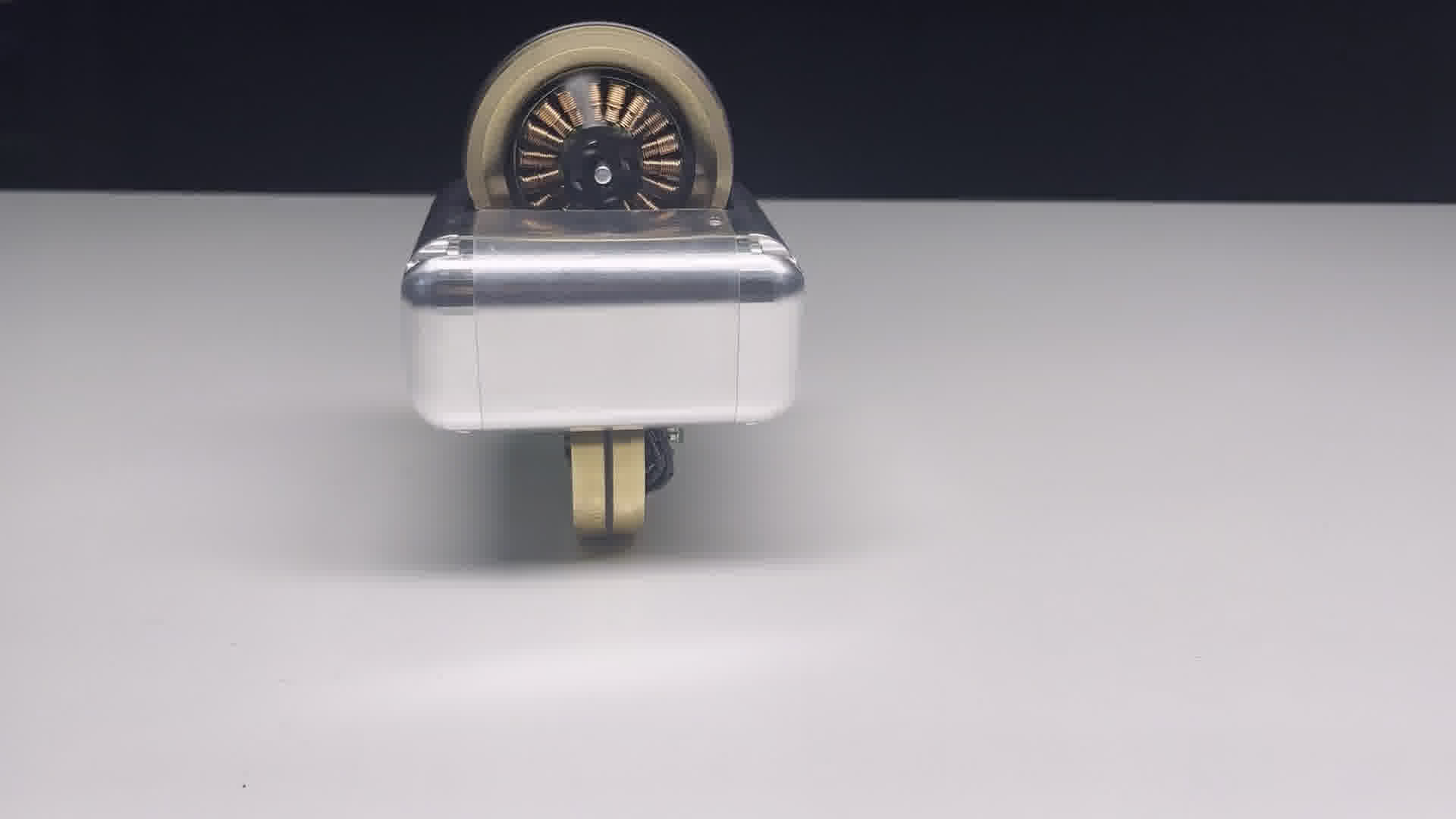}%
    \end{subfigure}
    \begin{subfigure}{1.8in}%
        \input{figures/wheelbot_standup/pitch/pitch.tex}%
        \subcaption{Pitch stand-up.}\label{fig:standup:pitch}
    \end{subfigure}%
    \begin{subfigure}{1.2in}%
        \input{figures/wheelbot_standup/roll/roll.tex}%
        \subcaption{Roll stand-up.}\label{fig:standup:roll}
    \end{subfigure}%
    \vspace{1em}
    \begin{subfigure}{3.2in}
        \includegraphics[height=0.533in,trim={5cm  3cm 35cm 8cm},clip]{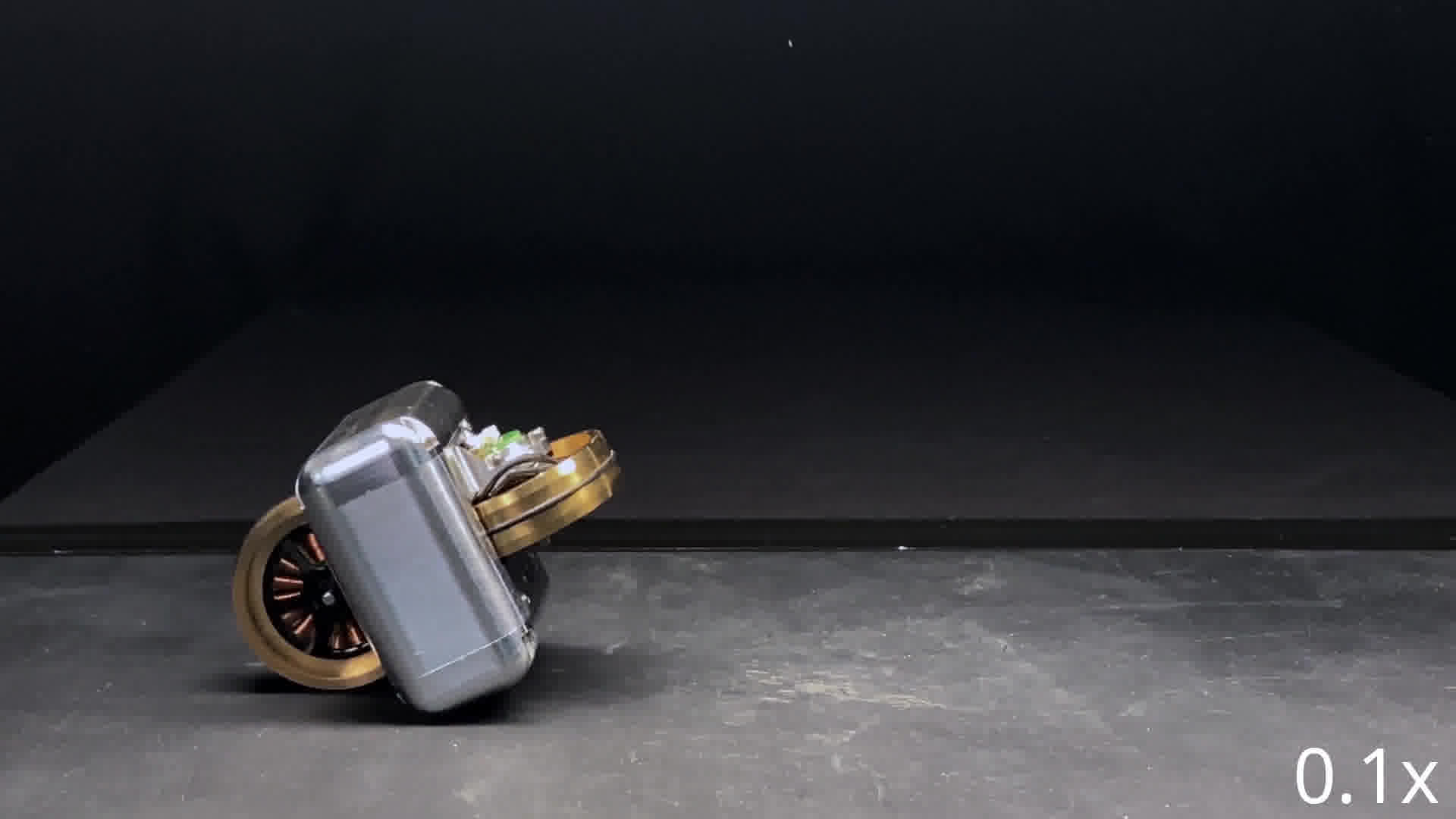}%
        \includegraphics[height=0.533in,trim={8cm  3cm 32cm 8cm},clip]{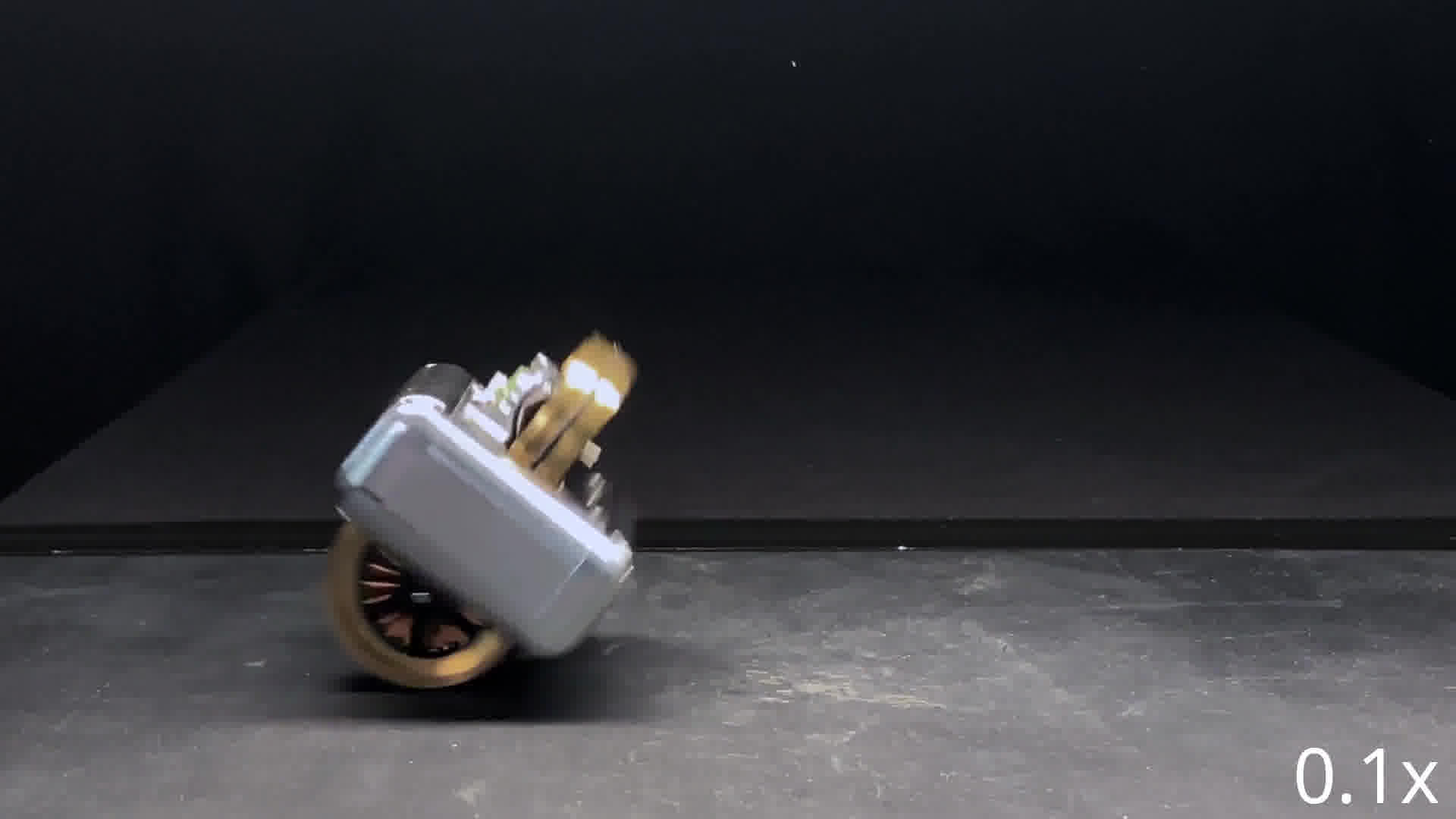}%
        \includegraphics[height=0.533in,trim={15cm 3cm 25cm 8cm},clip]{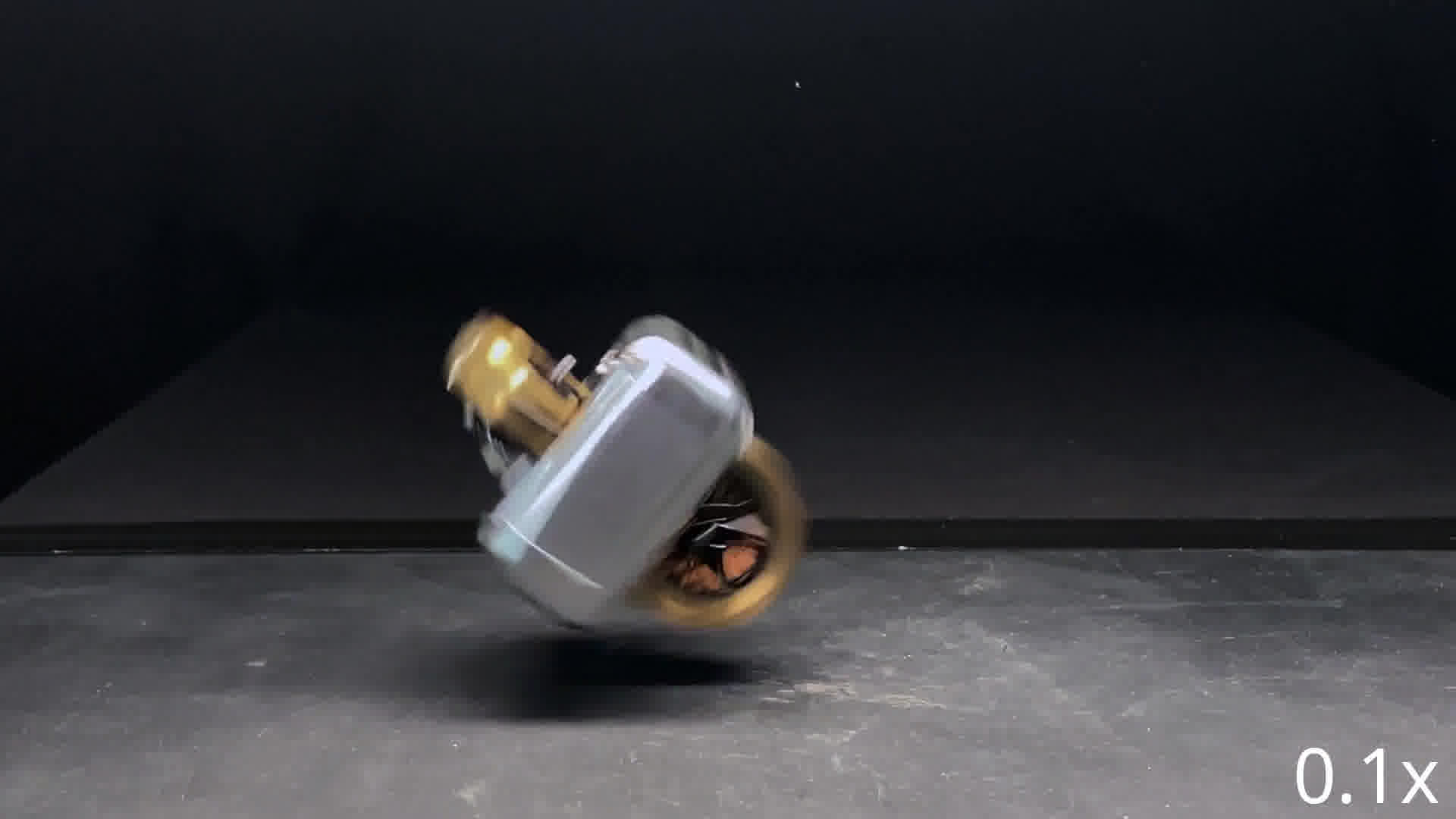}%
        \includegraphics[height=0.533in,trim={18cm 3cm 22cm 8cm},clip]{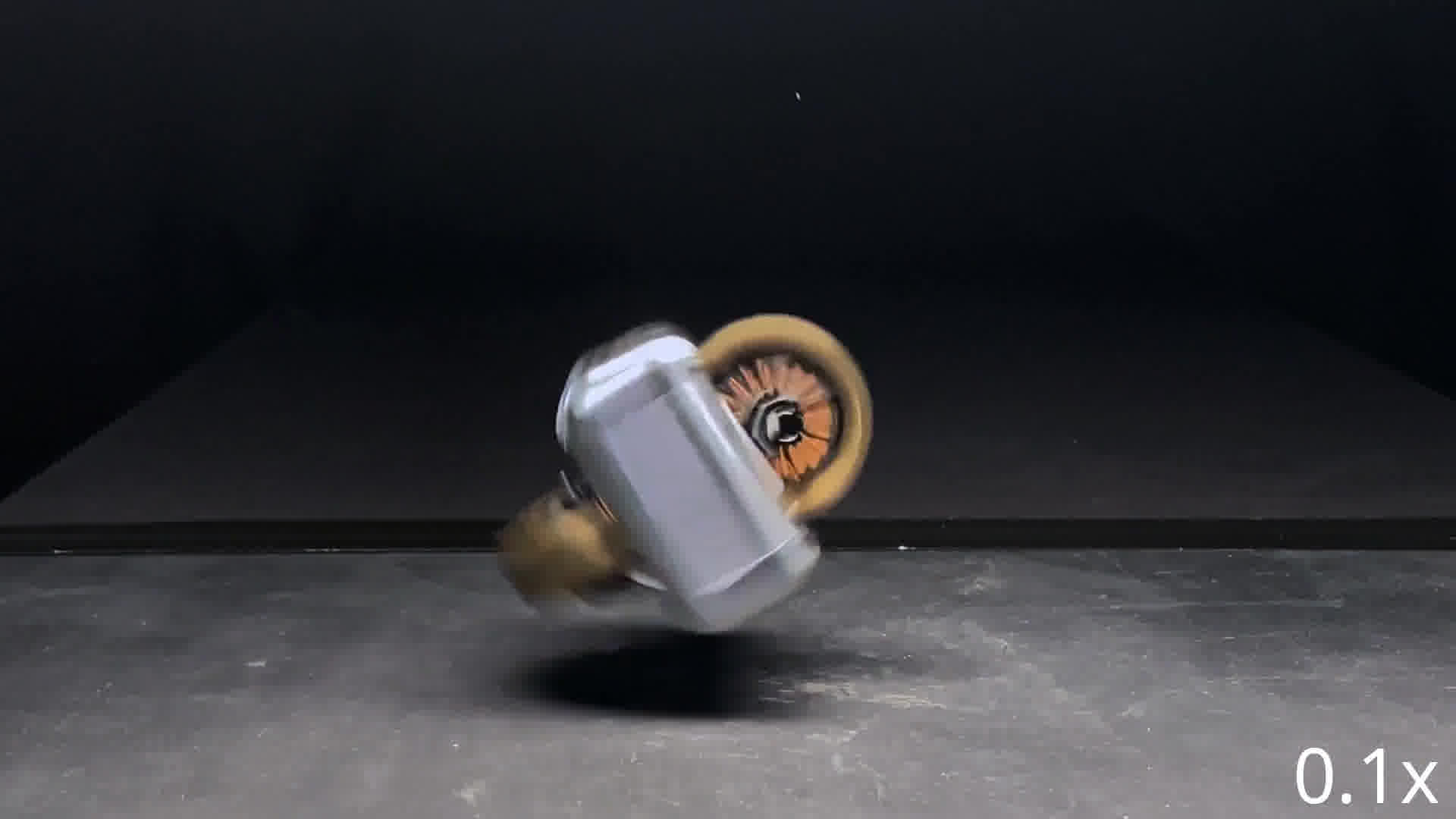}%
        \includegraphics[height=0.533in,trim={20cm 3cm 20cm 8cm},clip]{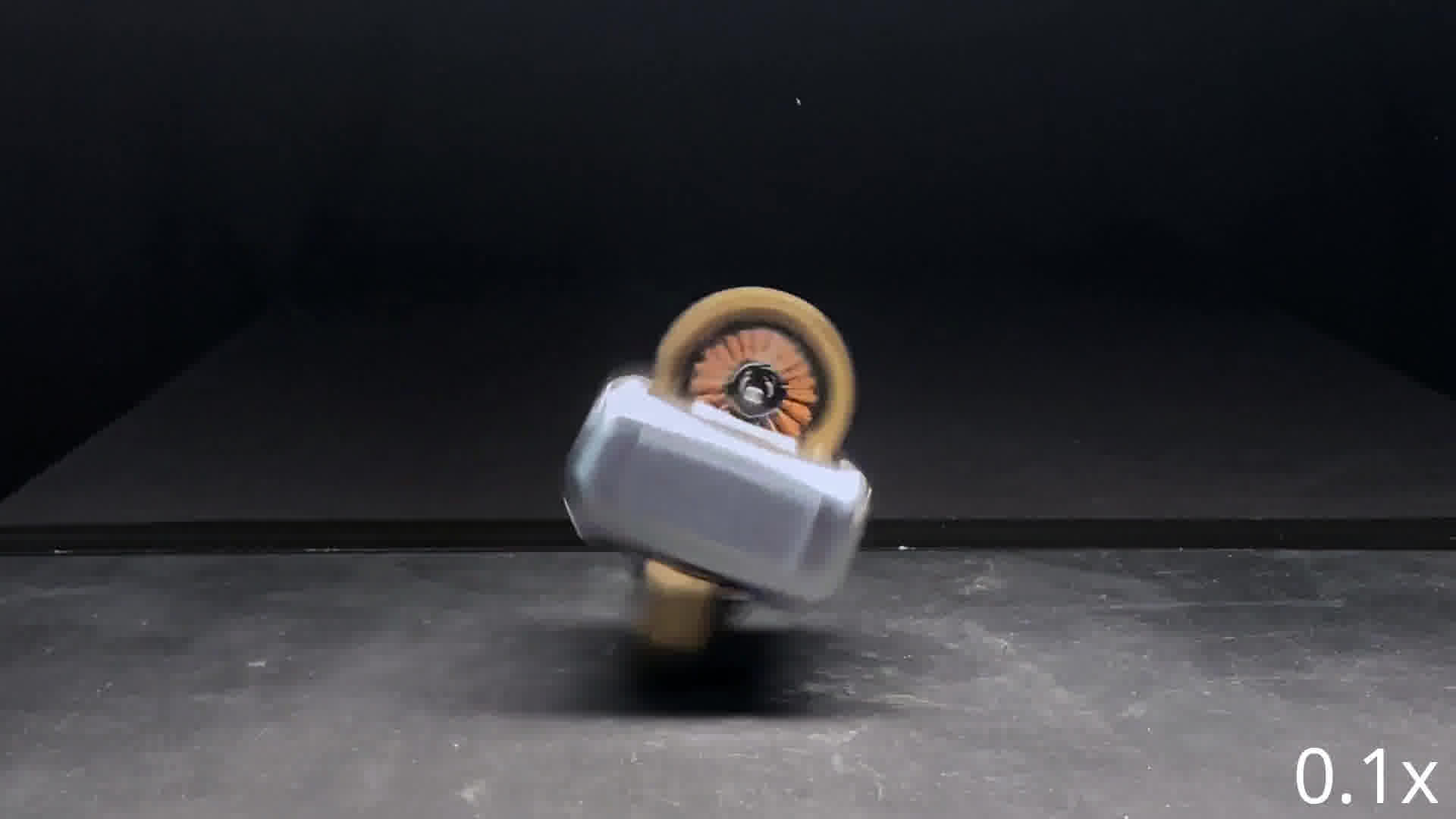}%
        \includegraphics[height=0.533in,trim={22cm 3cm 14cm 8cm},clip]{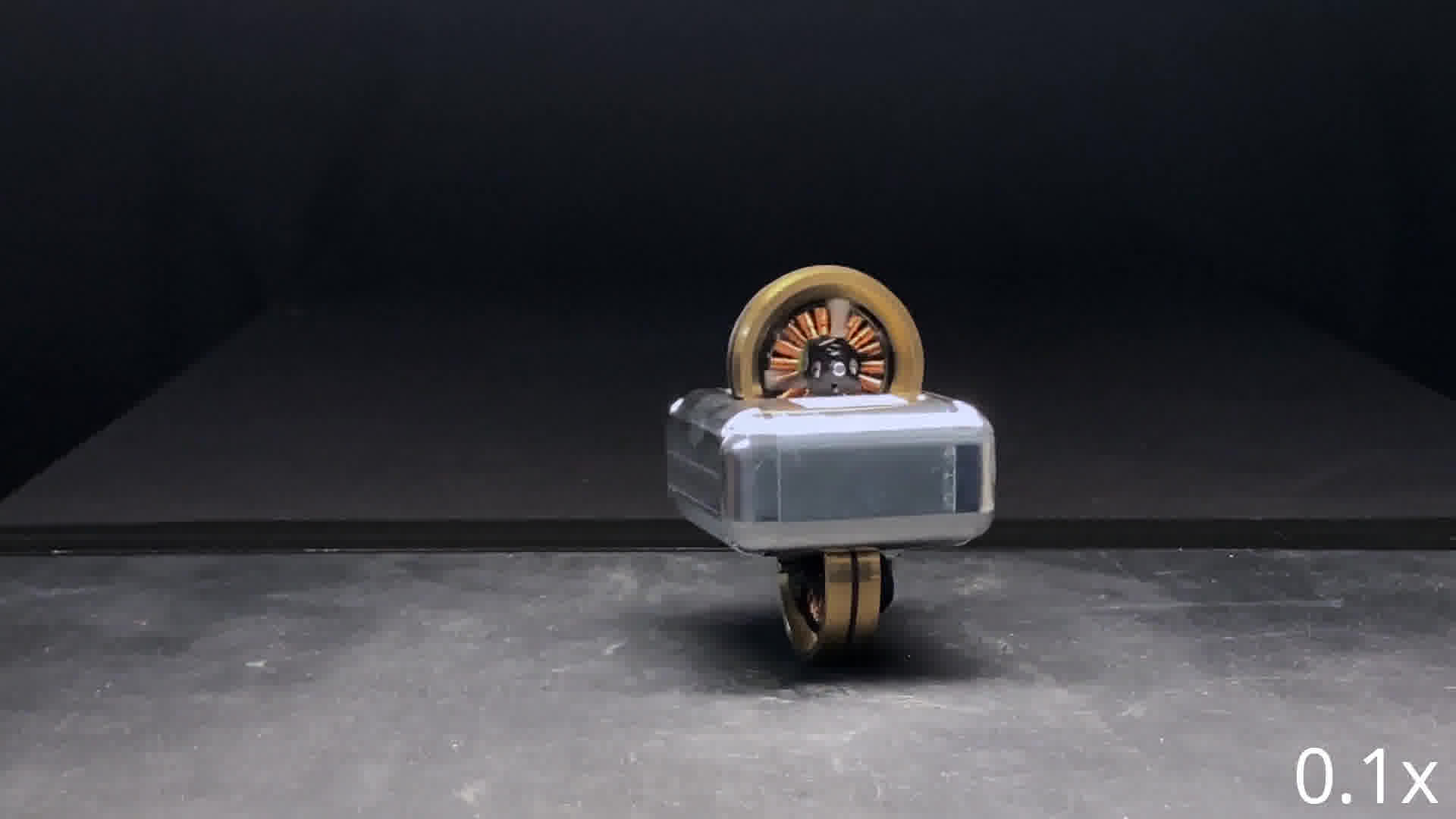}%
    \end{subfigure}
    \begin{subfigure}{3.2in}
        \input{figures/wheelbot_standup/flip/flip.tex}%
        \subcaption{Half flip.}\label{fig:standup:flip}
    \end{subfigure}
    \caption{Environment reset and half flip: The Wheelbot can stand up from any side by using its driving wheel (top left), its reaction wheel (top right), and even perform a flip stand-up (bottom). The maneuvers can be chained to reliably perform environment resets in episodic tasks.}
    \label{fig:standup}
\end{figure}

\label{sec:standup}
Similar to~\cite{geist2022wheelbot}, the new Mini Wheelbot can stand up in the pitch and roll direction (see Fig.~\ref{fig:standup:pitch} and ~\ref{fig:standup:roll}).
Standing up in the pitch direction involves one wheel quickly driving under the robot.
For the roll stand-up, the robot first spins up the reaction wheel before rapidly decelerating it thus excerting a counter torque that makes the robot stand up.
Both maneuvers are achieved via precomputed, open-loop command sequences before switching on the state-feedback controller (see Sec.~\ref{sec:statefeedback}) when entering~$\pm$\SI{30}{\degree} roll and pitch.

\subsection{Half Flip}
\label{sec:flip}
We implement a novel half-flip stand-up maneuver, visualized in Fig.~\ref{fig:standup:flip}.
This acrobatic maneuver pushes the Mini Wheelbot to its limits, illustrating the power of the hardware and hinting at future complex high-speed learning tasks.
The half flip is executed from an open loop action sequence taking about~\SI{160}{\milli\second} before switching on the balancing state-feedback controller for landing.
Similar to the pitch stand-up maneuver, the drive wheel quickly drives under the robot, however, with much higher torque, thus accelerating the Mini Wheelbot beyond the upright position.
After a~\SI{180}{\degree} rotation, the balancing position on the other wheel is reached, where the state-feedback controller starts balancing.

\section{Learning-based Control Experiments}
\label{sec:lbcexperiments}
The Mini Wheelbot involves a number of challenges, making it an interesting testbed for data-driven, learning-based control algorithms.
We showcase two of such algorithms.
In Sec.~\ref{sec:bo}, we use BO which has become popular for controller tuning in the last decade~\cite{chatzilygeroudis2019survey,paulson2023tutorial, marco2016automatic, calandra2016bayesian, berkenkamp2016safe} and allows us to automatically find excellent balancing controller gains in automated hardware experiments.
In Sec.~\ref{sec:ampc}, we use imitation learning from an expert MPC, also called approximate MPC (see~\cite{gonzalez2023neural} for a recent survey).
Approximate MPC avoids slow online optimization which enables sophisticated nonlinear MPC in fast feedback loops onboard robots~\cite{carius2020mpc,nubert2020safe} even on low-cost hardware~\cite{hose2024parameter}.

\subsection{Tuning Stand-up \& Balancing via Bayesian Optimization}
\label{sec:bo}
BO describes a family of black-box optimization algorithms that can be used for controller tuning based on a few interactions with the real world system~\cite{paulson2023tutorial, marco2016automatic, calandra2016bayesian, berkenkamp2016safe,chatzilygeroudis2019survey}. 
Instead of LQR in~\cite{geist2022wheelbot}, we use BO to tune the gains of the state-feedback controller~(\ref{eqn:feedbackgain}) in a direct, data-driven approach based on rewards collected in real-world experiments.
This is practically motivated: Tuning LQR cost matrices can be unintuitive and for some gain combinations, unmodeled high-frequency oscillations occur that are difficult to avoid through the choice of LQR cost.
In addition to finding excellent balancing controller gains, BO illustrates the advantage of automatic environment resets through the Mini Wheelbot's stand-up maneuvers for learning on episodic tasks.

\begin{figure}[tb]
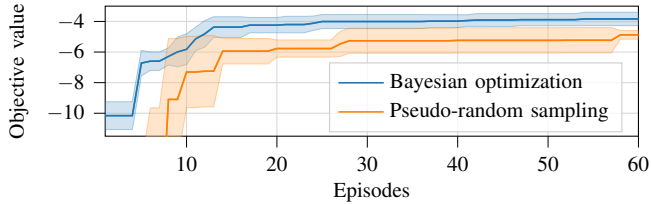

    \centering%
    \include{figures/bo_results/performance_improvement.tex}
    \caption{Performance of the best controller found by BO ({\color{blue}blue}) for a given number of trials (episodes) compared to a pseudo-random baseline ({\color{orange}orange}). Intervals are the standard deviation over five random seeds.}
    \label{fig:boperformanceimprovement}
\end{figure}

With BO, we find
\begin{align}\label{eqn:bo}
K^* = \argmax_{K} V(K)
\end{align}
based on (noisy) real-world evaluations of the true objective function~$V(K)$.
We define the objective function to be
\begin{align}
    \label{eqn:bo:cost}
    V(K) = 
        \begin{cases}
        -J_\text{c}, & \text{if crash}, \\
        -\frac{1}{T_\text{BO}}\sum_{t=0}^{T_\text{BO}} \|x(t)\|_{Q_\text{BO}}^2 - w_\text{vib}\cdot J_{\text{vib}}, & \text{else},
        \end{cases}
\end{align}
where~$T_\text{BO}$ is the experiments time horizon. The crash penalty~$J_\text{c}$ is empirically chosen slightly worse than a barely successfull experiment (e.g.,~$J_\text{c}=30$ for the experiment shown in Fig.~\ref{fig:boperformanceimprovement}).
For successful experiments, the objective value consists of the mean squared error over the state deviation from the equilibrium with diagonal weights~$Q_\text{BO}=\text{diag}([0,50,200,0,0.4,0.1,0,\dots]^\top)$.
The additional penalty~$J_\text{vib}$ weighted by~$w_\text{vib}=10^{-5}$ is
\begin{align}
    \label{eqn:bo:costvib}
    J_{vib} 
    = \sum_{t=1}^{T_\text{BO}} 
    ( 
        &|\Delta\dot{\phi}(t)|,
        \text{if}\; |\Delta\dot{\phi}(t)| \geq \alpha, \;
        \text{else}\; 0
    ),
\end{align}
with $\Delta\dot{\phi}(t)=\dot{\phi}(t)-\dot{\phi}(t-1)$, which effectively suppresses high frequency vibrations.

\begin{figure}[tb]
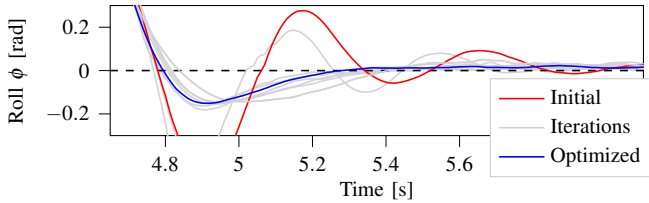

    \centering%
    \include{figures/bo_results/4d_roll_angle_comparison.tex}
    \caption{State evolution for a hand-tuned controller ({\color{red}red}), several episodes of BO ({\color{gray}gray}), and optimized controller ({\color{blue}blue}). The task is stabilizing after roll stand-up.}
    \label{fig:boclosedloop}
\end{figure}

For each episode of the optimization, the Wheelbot has to perform a stand-up maneuver and balance with a setpoint change in the drive-wheel position.
An excerpt of the closed loop trajectories from our experiments when the controller stabilizes the robot right after the stand-up is shown in Fig.~\ref{fig:boclosedloop}.
We use GPyTorch~\cite{gardner2018gpytorch} to solve (\ref{eqn:bo}).
For comparison, we provide a pseudo-random baseline based on space-filling Sobol sampling.
The performance improvement over the episodes for five random seeds is plotted in Fig.~\ref{fig:boperformanceimprovement}.
Even though, pseudo random sampling finds acceptable controllers eventually, BO has the better overall performance or requires less experiments with less failures.


\subsection{Fast Neural Network Approximate MPC}
\label{sec:ampc}
Despite great engineering efforts, nonlinear solvers still struggle to solve MPC optimization problems in real-time on embedded CPUs.
Instead, we use imitation learning to find an explicit mapping from states to actions in the form of a neural network that approximates the MPC described in Section~\ref{sec:nonlinearmpc}.
The approximate MPC is several orders of magnitude faster (onboard inference in less than~\SI{300}{\micro\second}) compared to solving the optimization problem~\eqref{eqn:mpc}, which takes several hundred milliseconds on a desktop CPU.
It thus allows a sophisticated nonlinear MPC with convoluted system dynamics, fine discretization, and a long prediction horizon to run onboard the robot.
With this, we achieve -- for the first time -- articulated driving following high level yaw and velocity setpoints, e.g., provided by keyboard.

\begin{figure}[tb]
    \newcommand{\photoheight}{1.2in}
    \newcommand{\plotwidth}{1.65in}
    \newcommand{\plotheight}{1.4in}
    \centering
    \begin{subfigure}[t]{1.6in}
        \centering
        \includegraphics[height=\photoheight,trim=0.5in 0 0.5in 0, clip]{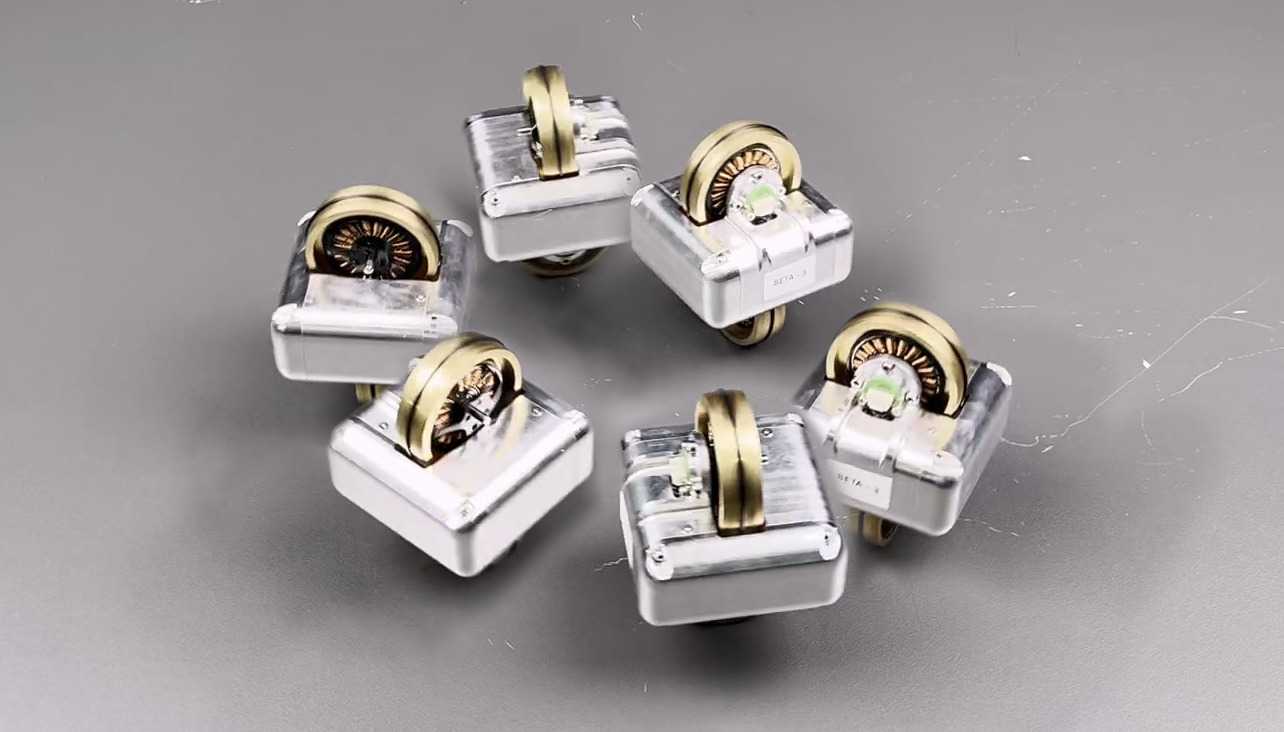}
    \end{subfigure}
    \begin{subfigure}[t]{1.6in}
        \centering
        \input{figures/yaw-control-plot/yaw_step_photo.tex}
    \end{subfigure}
    \par\medskip %
    \begin{subfigure}[t]{1.4in}
        \centering
        \begin{tikzpicture}[x=1in, y=1in,spy using outlines={circle, magnification=4, connect spies}]
  \useasboundingbox (-0.90, 0.1) rectangle (0.7, -0.85); %
    \definecolor{darkgray100}{RGB}{100,100,100}
    \definecolor{darkgray176}{RGB}{176,176,176}
    \definecolor{lightgray204}{RGB}{204,204,204}

    \begin{axis}[
        name=other,
        xshift=0mm,
        anchor=north,
        ytick distance = 0.2,
        xtick distance = 0.2,
        ytick style={color=black},
        ymajorgrids,
        ymin=-0.3, ymax=0.1,
        xmin=-0.2, xmax=0.2,
        y label style={yshift=-2mm},
        x label style={yshift=2mm},
        x grid style={darkgray176},
        xmajorgrids,
        xtick style={color=black},
        y grid style={darkgray176},
        ylabel={{Pos. y [m]}},
        xlabel={{Pos. x [m]}},
        label style={font=\scriptsize},
        ticklabel style={font=\scriptsize},
        height=\plotheight,
        legend style={
            fill opacity=1,
            draw opacity=1,
            text opacity=1,
            at={(1.02, -0.05)},
            anchor=south east,
            font=\scriptsize,
            draw=lightgray204
          },
        reverse legend,
        axis equal
        ]

    \addplot[thick, color=blue] table [x=x, y=y, col sep=comma] {figures/yaw-control-plot/yaw_tracking_error.csv};

    \end{axis}

    \end{tikzpicture}
    \end{subfigure}\hspace{0.2in}
    \begin{subfigure}[t]{1.6in}
        \begin{tikzpicture}[x=1in,y=1in]
  \useasboundingbox (-0.80, 0.1) rectangle (0.8, -0.85); %
    \definecolor{darkgray100}{RGB}{100,100,100}
    \definecolor{darkgray176}{RGB}{176,176,176}
    \definecolor{lightgray204}{RGB}{204,204,204}
    
        \begin{axis}[
            name=yaw,
            anchor=north,
            ytick style={color=black},
            ytick={-1.57, 0, 1.57},
            yticklabels={-$\frac{\pi}{2}$, $0$, $\frac{\pi}{2}$},
            ymajorgrids,
            ymin=-2, ymax=2,
            y label style={yshift=-2mm},
            x grid style={darkgray176},
            xmajorgrids,
            xmin=0, xmax=22,
            xtick style={color=black},
            y grid style={darkgray176},
            x label style={yshift=2mm},
            xlabel={Time [s]},
            ylabel={{Yaw $\psi$ [rad]}},
            height=\plotheight, width=\plotwidth,
            label style={font=\scriptsize},
            ticklabel style={font=\scriptsize},
            legend cell align={left},
            legend style={
              draw opacity=1,
              text opacity=1,
              at={(1.25,1.1)},
              anchor=north east,
              draw=lightgray204,
              font=\scriptsize
            },
            reverse legend,
        ]
    
        \addplot[mark=none, thick, color=darkgray100] table [x=time, y=yaw_setpoint, col sep=comma] {figures/yaw-control-plot/wheelbot_hardware_turbo_initial.csv};
        \addlegendentry{Reference};

        \addplot[mark=none, thick, color=blue] table [x=time, y=yaw, col sep=comma] {figures/yaw-control-plot/wheelbot_hardware_turbo_initial.csv};
        \addlegendentry{Measured};

        \end{axis}
    \end{tikzpicture}
    \end{subfigure}
    \caption{Approximate MPC experiments: Driving around based on keyboard heading and velocity commands (left) and yaw reference step response (right).
    The controller runs onboard the Mini Wheelbot.
    }
    \label{fig:mpcexperiments}
\end{figure}

We imitate the optimal solution to~(\ref{eqn:mpc}) by first sampling a large dataset of random states~$x$ and optimal solutions~$u^*(x)$ containing \num{3.5} million points.
We use a multi-layer perceptron with \num{4} layers, \num{100} neurons per layer, and a mixture of tangent hyperbolic and rectified linear activations as function approximator, which is trained in Jax.
To achieve fast control, we implement inference in C++ using Eigen onboard the Mini Wheelbot.
The final controller on the Mini Wheelbot hardware driving around based on keyboard heading and velocity commands is shown in Fig.~\ref{fig:mpcexperiments} and the supplementary video.
This is the first time that controlled driving along specified directions is achieved for a reaction wheel balancing unicycle of this kind.

\section{Conclusion}
We introduce the new Mini Wheelbot, a balancing, reaction wheel unicycle robot designed as a robust, compact, and powerful testbed for learning-based control.
We demonstrate the effectiveness of our platform in learning-based control tasks. First, we automatically tune the robot's state-feedback controller using real-world experiments via BO.
As BO requires repeated experiments, this demonstrates how the Mini Wheelbot's automatic environment reset facilitates learning-based control.
In a second illustrative application of learning-based control — imitation learning from an expert MPC — we show that sophisticated MPC schemes can be implemented without the burden of real-time optimization onboard the Mini Wheelbot.
With this approximate MPC, we achieve yaw control and articulated driving for the first time in this class of robots.
However, an in-depth theoretical analysis of this controller remains for future work.

During our experiments, we identified several open topics for future research.
First, the estimator could be enhanced to handle singularities in the orientation representation and saturation in the gyroscope.
Additionally, estimating contact could improve reliability of flip maneuvers and enable more acrobatic sequences with controlled contact switches.
Second, with articulated driving now feasible, we plan to use a fleet of Mini Wheelbots to test future algorithmic advances in multi-robot coordination. 
Finally, the Mini Wheelbot has proven to be a robust testbed for learning through repeated experiments and automatic environment resets.
As such, we envision benchmarking various learning algorithms, including reinforcement learning.

\section*{Acknowledgments}
We thank T.~Beyer, P.~Brunzema, D.~Buchholz, R.~Frohn,  R.~Geist, S.~Giedyk, J.~Menn, A.~Mitri, and M.~Ramirez.

\bibliographystyle{IEEEtran}
\bibliography{references}

\begin{thebibliography}{10}
\providecommand{\url}[1]{#1}
\csname url@samestyle\endcsname
\providecommand{\newblock}{\relax}
\providecommand{\bibinfo}[2]{#2}
\providecommand{\BIBentrySTDinterwordspacing}{\spaceskip=0pt\relax}
\providecommand{\BIBentryALTinterwordstretchfactor}{4}
\providecommand{\BIBentryALTinterwordspacing}{\spaceskip=\fontdimen2\font plus
\BIBentryALTinterwordstretchfactor\fontdimen3\font minus \fontdimen4\font\relax}
\providecommand{\BIBforeignlanguage}[2]{{%
\expandafter\ifx\csname l@#1\endcsname\relax
\typeout{** WARNING: IEEEtran.bst: No hyphenation pattern has been}%
\typeout{** loaded for the language `#1'. Using the pattern for}%
\typeout{** the default language instead.}%
\else
\language=\csname l@#1\endcsname
\fi
#2}}
\providecommand{\BIBdecl}{\relax}
\BIBdecl

\bibitem{geist2022wheelbot}
A.~R. Geist, J.~Fiene, N.~Tashiro, Z.~Jia, and S.~Trimpe, ``The {Wheelbot}: A jumping reaction wheel unicycle,'' \emph{IEEE Robotics and Automation Letters}, vol.~7, no.~4, pp. 9683--9690, 2022.

\bibitem{hofer2023one}
M.~Hofer, M.~Muehlebach, and R.~D’Andrea, ``The one-wheel {Cubli}: A 3d inverted pendulum that can balance with a single reaction wheel,'' \emph{Mechatronics}, vol.~91, p. 102965, 2023.

\bibitem{carron2023chronos}
A.~Carron, S.~Bodmer, L.~Vogel, R.~Zurbr{\"u}gg, D.~Helm, R.~Rickenbach, S.~Muntwiler, J.~Sieber, and M.~N. Zeilinger, ``Chronos and {CRS}: Design of a miniature car-like robot and a software framework for single and multi-agent robotics and control,'' in \emph{2023 IEEE International Conference on Robotics and Automation (ICRA)}.\hskip 1em plus 0.5em minus 0.4em\relax IEEE, 2023, pp. 1371--1378.

\bibitem{bodmer2024optimization}
S.~Bodmer, L.~Vogel, S.~Muntwiler, A.~Hansson, T.~Bodewig, J.~Wahlen, M.~N. Zeilinger, and A.~Carron, ``Optimization-based system identification and moving horizon estimation using low-cost sensors for a miniature car-like robot,'' \emph{arXiv preprint arXiv:2404.08362}, 2024.

\bibitem{o2020f1tenth}
M.~O'Kelly, H.~Zheng, D.~Karthik, and R.~Mangharam, ``{F1tenth}: An open-source evaluation environment for continuous control and reinforcement learning,'' \emph{Proceedings of Machine Learning Research}, vol. 123, 2020.

\bibitem{hanover2024autonomous}
D.~Hanover, A.~Loquercio, L.~Bauersfeld, A.~Romero, R.~Penicka, Y.~Song, G.~Cioffi, E.~Kaufmann, and D.~Scaramuzza, ``Autonomous drone racing: A survey,'' \emph{IEEE Transactions on Robotics}, 2024.

\bibitem{kim2017design}
S.~Kim, P.~M. Wensing \emph{et~al.}, ``Design of dynamic legged robots,'' \emph{Foundations and Trends{\textregistered} in Robotics}, vol.~5, no.~2, pp. 117--190, 2017.

\bibitem{mondada1994mobile}
F.~Mondada, E.~Franzi, and P.~Ienne, ``Mobile robot miniaturisation: A tool for investigation in control algorithms,'' in \emph{Experimental Robotics III: The 3rd International Symposium, Kyoto, Japan, October 28--30, 1993}.\hskip 1em plus 0.5em minus 0.4em\relax Springer, 1994, pp. 501--513.

\bibitem{johnson2006mobile}
D.~Johnson, T.~Stack, R.~Fish, D.~M. Flickinger, L.~Stoller, R.~Ricci, and J.~Lepreau, ``Mobile emulab: A robotic wireless and sensor network testbed,'' in \emph{Proceedings IEEE INFOCOM 2006. 25TH IEEE International Conference on Computer Communications}.\hskip 1em plus 0.5em minus 0.4em\relax IEEE, 2006, pp. 1--12.

\bibitem{mondada2009puck}
F.~Mondada, M.~Bonani, X.~Raemy, J.~Pugh, C.~Cianci, A.~Klaptocz, S.~Magnenat, J.-C. Zufferey, D.~Floreano, and A.~Martinoli, ``The e-puck, a robot designed for education in engineering,'' in \emph{Proceedings of the 9th conference on autonomous robot systems and competitions}, vol.~1, no.~1.\hskip 1em plus 0.5em minus 0.4em\relax IPCB: Instituto Polit{\'e}cnico de Castelo Branco, 2009, pp. 59--65.

\bibitem{rubenstein2015aerobot}
M.~Rubenstein, B.~Cimino, R.~Nagpal, and J.~Werfel, ``Aerobot: An affordable one-robot-per-student system for early robotics education,'' in \emph{2015 IEEE International Conference on Robotics and Automation (ICRA)}.\hskip 1em plus 0.5em minus 0.4em\relax IEEE, 2015, pp. 6107--6113.

\bibitem{pickem2015gritsbot}
D.~Pickem, M.~Lee, and M.~Egerstedt, ``The gritsbot in its natural habitat-a multi-robot testbed,'' in \emph{2015 IEEE International conference on robotics and automation (ICRA)}.\hskip 1em plus 0.5em minus 0.4em\relax IEEE, 2015, pp. 4062--4067.

\bibitem{wilson2016pheeno}
S.~Wilson, R.~Gameros, M.~Sheely, M.~Lin, K.~Dover, R.~Gevorkyan, M.~Haberland, A.~Bertozzi, and S.~Berman, ``Pheeno, a versatile swarm robotic research and education platform,'' \emph{IEEE Robotics and Automation Letters}, vol.~1, no.~2, pp. 884--891, 2016.

\bibitem{paull2017duckietown}
L.~Paull, J.~Tani, H.~Ahn, J.~Alonso-Mora, L.~Carlone, M.~Cap, Y.~F. Chen, C.~Choi, J.~Dusek, Y.~Fang \emph{et~al.}, ``Duckietown: an open, inexpensive and flexible platform for autonomy education and research,'' in \emph{2017 IEEE International Conference on Robotics and Automation (ICRA)}.\hskip 1em plus 0.5em minus 0.4em\relax IEEE, 2017, pp. 1497--1504.

\bibitem{hsieh2006economical}
C.~H. Hsieh, Y.-L. Chuang, Y.~Huang, K.~K. Leung, A.~L. Bertozzi, and E.~Frazzoli, ``An economical micro-car testbed for validation of cooperative control strategies,'' in \emph{2006 American Control Conference}.\hskip 1em plus 0.5em minus 0.4em\relax IEEE, 2006, pp. 6--pp.

\bibitem{liniger2015optimization}
A.~Liniger, A.~Domahidi, and M.~Morari, ``Optimization-based autonomous racing of 1: 43 scale {RC} cars,'' \emph{Optimal Control Applications and Methods}, vol.~36, no.~5, pp. 628--647, 2015.

\bibitem{goldfain2019autorally}
B.~Goldfain, P.~Drews, C.~You, M.~Barulic, O.~Velev, P.~Tsiotras, and J.~M. Rehg, ``Autorally: An open platform for aggressive autonomous driving,'' \emph{IEEE Control Systems Magazine}, vol.~39, no.~1, pp. 26--55, 2019.

\bibitem{gonzales2018planning}
J.~M. Gonzales, \emph{Planning and control of drift maneuvers with the Berkeley autonomous race car}.\hskip 1em plus 0.5em minus 0.4em\relax University of California, Berkeley, 2018.

\bibitem{gajamohan2012cubli}
M.~Gajamohan, M.~Merz, I.~Thommen, and R.~D'Andrea, ``The {Cubli}: A cube that can jump up and balance,'' in \emph{2012 IEEE/RSJ International Conference on Intelligent Robots and Systems}.\hskip 1em plus 0.5em minus 0.4em\relax IEEE, 2012, pp. 3722--3727.

\bibitem{mayr2015mechatronic}
J.~Mayr, F.~Spanlang, and H.~Gattringer, ``Mechatronic design of a self-balancing three-dimensional inertia wheel pendulum,'' \emph{Mechatronics}, vol.~30, pp. 1--10, 2015.

\bibitem{muehlebach2016nonlinear}
M.~Muehlebach and R.~D’Andrea, ``Nonlinear analysis and control of a reaction wheel-based 3-d inverted pendulum,'' \emph{IEEE Transactions on Control Systems Technology}, vol.~25, no.~1, pp. 235--246, 2016.

\bibitem{klemm2019ascento}
V.~Klemm, A.~Morra, C.~Salzmann, F.~Tschopp, K.~Bodie, L.~Gulich, N.~K{\"u}ng, D.~Mannhart, C.~Pfister, M.~Vierneisel \emph{et~al.}, ``Ascento: A two-wheeled jumping robot,'' in \emph{2019 International Conference on Robotics and Automation (ICRA)}.\hskip 1em plus 0.5em minus 0.4em\relax IEEE, 2019, pp. 7515--7521.

\bibitem{nagarajan2014ballbot}
U.~Nagarajan, G.~Kantor, and R.~Hollis, ``The {Ballbot}: An omnidirectional balancing mobile robot,'' \emph{The International Journal of Robotics Research}, vol.~33, no.~6, pp. 917--930, 2014.

\bibitem{pickem2017robotarium}
D.~Pickem, P.~Glotfelter, L.~Wang, M.~Mote, A.~Ames, E.~Feron, and M.~Egerstedt, ``The robotarium: A remotely accessible swarm robotics research testbed,'' in \emph{2017 IEEE International Conference on Robotics and Automation (ICRA)}.\hskip 1em plus 0.5em minus 0.4em\relax IEEE, 2017, pp. 1699--1706.

\bibitem{schwab2020experimental}
A.~Schwab, L.-M. Reichelt, P.~Welz, and J.~Lunze, ``Experimental evaluation of an adaptive cruise control and cooperative merging concept,'' in \emph{2020 IEEE Conference on Control Technology and Applications (CCTA)}.\hskip 1em plus 0.5em minus 0.4em\relax IEEE, 2020, pp. 318--325.

\bibitem{trimpe2012balancing}
S.~Trimpe and R.~D’Andrea, ``The balancing cube: A dynamic sculpture as test bed for distributed estimation and control,'' \emph{IEEE Control Systems Magazine}, vol.~32, no.~6, pp. 48--75, 2012.

\bibitem{schoonwinkel1988design}
A.~Schoonwinkel, \emph{Design and test of a computer-stabilized unicycle}.\hskip 1em plus 0.5em minus 0.4em\relax Stanford University, 1988.

\bibitem{vos1990dynamics}
D.~W. Vos and A.~H. Von~Flotow, ``Dynamics and nonlinear adaptive control of an autonomous unicycle: Theory and experiment,'' in \emph{29th IEEE Conference on Decision and Control}.\hskip 1em plus 0.5em minus 0.4em\relax IEEE, 1990, pp. 182--187.

\bibitem{xu2011pendulum}
J.-X. Xu, A.~Al~Mamun, and Y.~Daud, ``Pendulum-balanced autonomous unicycle: Conceptual design and dynamics model,'' in \emph{2011 ieee 5th international conference on robotics, automation and mechatronics (ram)}.\hskip 1em plus 0.5em minus 0.4em\relax IEEE, 2011, pp. 51--56.

\bibitem{daud2017dynamic}
Y.~Daud, A.~Al~Mamun, and J.-X. Xu, ``Dynamic modeling and characteristics analysis of lateral-pendulum unicycle robot,'' \emph{Robotica}, vol.~35, no.~3, pp. 537--568, 2017.

\bibitem{lee2012decoupled}
J.~Lee, S.~Han, and J.~Lee, ``Decoupled dynamic control for pitch and roll axes of the unicycle robot,'' \emph{IEEE Transactions on Industrial Electronics}, vol.~60, no.~9, pp. 3814--3822, 2012.

\bibitem{jae2011fuzzy}
L.~Jae-Oh, H.~In-Woo, and L.~Jang-Myung, ``Fuzzy sliding mode control of unicycle robot,'' in \emph{2011 8th international conference on ubiquitous robots and ambient intelligence (urai)}.\hskip 1em plus 0.5em minus 0.4em\relax IEEE, 2011, pp. 521--524.

\bibitem{li2012attitude}
Y.~Li, J.-O. Lee, and J.~Lee, ``Attitude control of the unicycle robot using fuzzy-sliding mode control,'' in \emph{Intelligent Robotics and Applications: 5th International Conference, ICIRA 2012, Montreal, QC, Canada, October 3-5, 2012, Proceedings, Part III 5}.\hskip 1em plus 0.5em minus 0.4em\relax Springer, 2012, pp. 62--72.

\bibitem{rosyidi2016speed}
M.~A. Rosyidi, E.~H. Binugroho, S.~E.~R. Charel, R.~S. Dewanto, and D.~Pramadihanto, ``Speed and balancing control for unicycle robot,'' in \emph{2016 International Electronics Symposium (IES)}.\hskip 1em plus 0.5em minus 0.4em\relax IEEE, 2016, pp. 19--24.

\bibitem{neves2021discrete}
G.~P. Neves and B.~A. Ang{\'e}lico, ``A discrete lqr applied to a self-balancing reaction wheel unicycle: Modeling, construction and control,'' in \emph{2021 American control conference (ACC)}.\hskip 1em plus 0.5em minus 0.4em\relax IEEE, 2021, pp. 777--782.

\bibitem{rizal2015point}
Y.~Rizal, C.-T. Ke, and M.-T. Ho, ``Point-to-point motion control of a unicycle robot: Design, implementation, and validation,'' in \emph{2015 IEEE International Conference on Robotics and Automation (ICRA)}.\hskip 1em plus 0.5em minus 0.4em\relax IEEE, 2015, pp. 4379--4384.

\bibitem{jin2010balancing}
H.~Jin, J.~Hwang, and J.~Lee, ``A balancing control strategy for a one-wheel pendulum robot based on dynamic model decomposition: Simulations and experiments,'' \emph{IEEE/ASME Transactions on Mechatronics}, vol.~16, no.~4, pp. 763--768, 2010.

\bibitem{lehmann2021micro}
\BIBentryALTinterwordspacing
R.~Lehmann, Se-Bi, N.~Hauser, and C.~Lutz, ``{µMotor}: A motor controller for {BLDC} and {DC} motors up to {250W}.'' [Online]. Available: \url{https://github.com/roboterclubaachen/micro-motor}
\BIBentrySTDinterwordspacing

\bibitem{piccoli2016anticogging}
M.~Piccoli and M.~Yim, ``Anticogging: Torque ripple suppression, modeling, and parameter selection,'' \emph{The international journal of robotics research}, vol.~35, no. 1-3, pp. 148--160, 2016.

\bibitem{trimpe2010accelerometer}
S.~Trimpe and R.~D'Andrea, ``Accelerometer-based tilt estimation of a rigid body with only rotational degrees of freedom,'' in \emph{2010 IEEE International Conference on Robotics and Automation}.\hskip 1em plus 0.5em minus 0.4em\relax IEEE, 2010, pp. 2630--2636.

\bibitem{frey2023fast}
J.~Frey, J.~De~Schutter, and M.~Diehl, ``Fast integrators with sensitivity propagation for use in {CasADi},'' in \emph{2023 European Control Conference (ECC)}.\hskip 1em plus 0.5em minus 0.4em\relax IEEE, 2023.

\bibitem{bock1983recent}
H.~G. Bock, ``Recent advances in parameteridentification techniques for {ODE},'' in \emph{Numerical Treatment of Inverse Problems in Differential and Integral Equations: Proceedings of an International Workshop, Heidelberg, Fed. Rep. of Germany, August 30—September 3, 1982}.\hskip 1em plus 0.5em minus 0.4em\relax Springer, 1983, pp. 95--121.

\bibitem{valluru2017development}
J.~Valluru, P.~Lakhmani, S.~C. Patwardhan, and L.~T. Biegler, ``Development of moving window state and parameter estimators under maximum likelihood and bayesian frameworks,'' \emph{Journal of Process Control}, vol.~60, pp. 48--67, 2017.

\bibitem{simpson2023efficient}
L.~Simpson, A.~Ghezzi, J.~Asprion, and M.~Diehl, ``An efficient method for the joint estimation of system parameters and noise covariances for linear time-variant systems,'' in \emph{2023 62nd IEEE Conference on Decision and Control (CDC)}.\hskip 1em plus 0.5em minus 0.4em\relax IEEE, 2023, pp. 4524--4529.

\bibitem{andersson2019casadi}
J.~A. Andersson, J.~Gillis, G.~Horn, J.~B. Rawlings, and M.~Diehl, ``Casadi: a software framework for nonlinear optimization and optimal control,'' \emph{Mathematical Programming Computation}, vol.~11, pp. 1--36, 2019.

\bibitem{wachter2006implementation}
A.~W{\"a}chter and L.~T. Biegler, ``On the implementation of an interior-point filter line-search algorithm for large-scale nonlinear programming,'' \emph{Mathematical programming}, vol. 106, pp. 25--57, 2006.

\bibitem{rawlings2017model}
J.~B. Rawlings, D.~Q. Mayne, M.~Diehl \emph{et~al.}, \emph{Model predictive control: Theory, computation, and design}.\hskip 1em plus 0.5em minus 0.4em\relax Nob Hill Publishing Madison, WI, 2017, vol.~2.

\bibitem{chatzilygeroudis2019survey}
K.~Chatzilygeroudis, V.~Vassiliades, F.~Stulp, S.~Calinon, and J.-B. Mouret, ``A survey on policy search algorithms for learning robot controllers in a handful of trials,'' \emph{IEEE Transactions on Robotics}, vol.~36, no.~2, pp. 328--347, 2019.

\bibitem{paulson2023tutorial}
J.~A. Paulson, F.~Sorourifar, and A.~Mesbah, ``A tutorial on derivative-free policy learning methods for interpretable controller representations,'' in \emph{2023 American Control Conference (ACC)}, 2023, pp. 1295--1306.

\bibitem{marco2016automatic}
A.~Marco, P.~Hennig, J.~Bohg, S.~Schaal, and S.~Trimpe, ``Automatic {LQR} tuning based on {Gaussian} process global optimization,'' in \emph{2016 IEEE international conference on robotics and automation (ICRA)}.\hskip 1em plus 0.5em minus 0.4em\relax IEEE, 2016, pp. 270--277.

\bibitem{calandra2016bayesian}
R.~Calandra, A.~Seyfarth, J.~Peters, and M.~P. Deisenroth, ``{Bayesian} optimization for learning gaits under uncertainty: An experimental comparison on a dynamic bipedal walker,'' \emph{Annals of Mathematics and Artificial Intelligence}, vol.~76, pp. 5--23, 2016.

\bibitem{berkenkamp2016safe}
F.~Berkenkamp, A.~P. Schoellig, and A.~Krause, ``Safe controller optimization for quadrotors with gaussian processes,'' in \emph{2016 IEEE international conference on robotics and automation (ICRA)}.\hskip 1em plus 0.5em minus 0.4em\relax IEEE, 2016, pp. 491--496.

\bibitem{gonzalez2023neural}
C.~Gonzalez, H.~Asadi, L.~Kooijman, and C.~P. Lim, ``Neural networks for fast optimisation in model predictive control: a review,'' \emph{arXiv preprint arXiv:2309.02668}, 2023.

\bibitem{carius2020mpc}
J.~Carius, F.~Farshidian, and M.~Hutter, ``{MPC}-{N}et: A first principles guided policy search,'' \emph{IEEE Robotics and Automation Letters}, vol.~5, no.~2, pp. 2897--2904, 2020.

\bibitem{nubert2020safe}
J.~Nubert, J.~K{\"o}hler, V.~Berenz, F.~Allg{\"o}wer, and S.~Trimpe, ``Safe and fast tracking on a robot manipulator: Robust {MPC} and neural network control,'' \emph{IEEE Robotics and Automation Letters}, vol.~5, no.~2, pp. 3050--3057, 2020.

\bibitem{hose2024parameter}
H.~Hose, A.~Gr\"{a}fe, and S.~Trimpe, ``Parameter-adaptive approximate {MPC}: {T}uning neural-network controllers without retraining,'' in \emph{Proceedings of the 6th Annual Learning for Dynamics and Control Conference}, ser. Proceedings of Machine Learning Research, A.~Abate, M.~Cannon, K.~Margellos, and A.~Papachristodoulou, Eds., vol. 242.\hskip 1em plus 0.5em minus 0.4em\relax PMLR, 15--17 Jul 2024, pp. 349--360.

\bibitem{gardner2018gpytorch}
J.~Gardner, G.~Pleiss, K.~Q. Weinberger, D.~Bindel, and A.~G. Wilson, ``Gpytorch: Blackbox matrix-matrix gaussian process inference with {GPU} acceleration,'' \emph{Advances in neural information processing systems}, vol.~31, 2018.

\end{thebibliography}

\end{document}